%% file: 0_main.tex
\documentclass{article}

\PassOptionsToPackage{numbers,sort,compress}{natbib}

\usepackage[final]{neurips_2025}

\input{preamble}

\usepackage[utf8]{inputenc} 
\usepackage[T1]{fontenc}    
\usepackage{hyperref}
\usepackage[capitalise]{cleveref}

\usepackage{url}            
\usepackage{booktabs}       
\usepackage{amsfonts}       
\usepackage{nicefrac}       
\usepackage{microtype}      
\usepackage{xcolor}         

\title{\ours: Unifying Channel-Aware Masked Autoencoders and Multi-Channel Vision Transformers  for Improved Cross-Channel Learning}

\author{
Chau Pham\\
Boston University\\
\texttt{chaupham@bu.edu}
\And
Juan C. Caicedo\\
Morgridge Institute, UW–Madison\\
\texttt{juan.caicedo@wisc.edu}
\And
Bryan A. Plummer\\
Boston University\\
\texttt{bplum@bu.edu}
}

\begin{document}
\maketitle

\input{1_abstract}

\input{2_introduction}

\input{3_related_work}

\input{4_method}

\input{5_experiments}

\input{6_analysis_and_discussion}

\input{7_conclusion}

{\small
\bibliographystyle{unsrtnat}
\bibliography{my_bib}
}

\newpage
\appendix
\input{8_appendix}

\end{document}

%% file: preamble.tex
\usepackage{multirow}
\usepackage{epsfig}
\usepackage{graphicx}
\usepackage{subcaption}
\usepackage{amssymb}
\usepackage{booktabs}
\usepackage{xspace}
\usepackage{enumitem}
\usepackage{tikz} 
\usepackage{icomma} 
\usepackage[linesnumbered,ruled,vlined]{algorithm2e} 
\usepackage{algorithmic}
\usepackage{amsmath}

\SetKwInOut{Input}{Input}
\SetKwInOut{Output}{Output}

\makeatletter
\DeclareRobustCommand\onedot{\futurelet\@let@token\@onedot}
\def\@onedot{\ifx\@let@token.\else.\null\fi\xspace}
\def\eg{\emph{e.g}\onedot} 
\def\ie{\emph{i.e}\onedot}

\makeatother

\def\newmethodfont{\fontfamily{lmr}}
\newcommand{\methodfont}[1]{{\fontfamily{lmr}\selectfont #1}}

\newcommand{\depthwisevit}{{\newmethodfont\selectfont DepthwiseViT}}
\newcommand{\templatevit}{{\newmethodfont\selectfont TemplateMixingViT}}
\newcommand{\hypervit}{{\newmethodfont\selectfont HyperNetViT}}

\newcommand{\vit}{{\newmethodfont\selectfont ViT}}
\newcommand{\chadavit}{{\newmethodfont\selectfont ChAda-ViT}}
\newcommand{\channelvit}{{\newmethodfont\selectfont ChannelViT}}
\newcommand{\dichavit}{{\newmethodfont\selectfont DiChaViT}}

\newcommand{\simclr}{{\newmethodfont\selectfont SimCLR}}
\newcommand{\simsiam}{{\newmethodfont\selectfont SimSiam}}
\newcommand{\dino}{{\newmethodfont\selectfont DINO}}
\newcommand{\dinotwo}{{\newmethodfont\selectfont DINOv2}}
\newcommand{\ibot}{{\newmethodfont\selectfont iBOT}}

\newcommand{\camae}{{\newmethodfont\selectfont CA-MAE}}

\newcommand{\ours}{{\newmethodfont\selectfont ChA-MAEViT}}
\newcommand{\oursbr}{\ours{} (ours)}
\def\oursfullnamewithabbre{\textbf{Ch}annel-\textbf{A}ware \textbf{MAE} $-$ multichannel \textbf{Vi}sion \textbf{T}ransformer (\ours)}

\newcommand{\ourmask}{{Dynamic Channel-Patch Masking}}
\newcommand{\ourmaskwithfont}{{\newmethodfont\selectfont \ourmask}}
\newcommand{\ourmaskwithfontandabbr}{{\newmethodfont\selectfont Dynamic Channel-Patch (DCP) Masking}}
\newcommand{\ourmaskwithabbr}{{Dynamic Channel-Patch (DCP) Masking}}
\newcommand{\ourmaskabbr}{DCP Masking}

\def\ourimprove{$3.0-21.5\%$}

\usepackage{tikz}

\definecolor{pinkcolor2}{RGB}{252,66,162}
\definecolor{darkgreen}{RGB}{55, 118, 28}
\definecolor{redcolor}{RGB}{255,0,0}
\definecolor{bluecolor}{RGB}{0,112,192}
\definecolor{orangecolor}{RGB}{255, 147, 0}
\definecolor{purplecolor}{RGB}{160, 43, 147}


%% file: 1_abstract.tex
\begin{abstract}
Prior work using Masked Autoencoders (MAEs) typically relies on random patch masking based on the assumption that images have significant redundancies across different channels, allowing for the reconstruction of masked content using cross-channel correlations. However, this assumption does not hold in Multi-Channel Imaging (MCI), where channels may provide complementary information with minimal feature overlap. Thus, these MAEs primarily learn local structures within individual channels from patch reconstruction, failing to fully leverage cross-channel interactions and limiting their MCI effectiveness.
In this paper, we present ChA-MAEViT, an MAE-based method that enhances feature learning across MCI channels via four key strategies: (1) dynamic channel-patch masking, which compels the model to reconstruct missing channels in addition to masked patches, thereby enhancing cross-channel dependencies and improving robustness to varying channel configurations; (2) memory tokens, which serve as long-term memory aids to promote information sharing across channels, addressing the challenges of reconstructing structurally diverse channels; (3) hybrid token fusion module, which merges fine-grained patch tokens with a global class token to capture richer representations; and (4) Channel-Aware Decoder, a lightweight decoder utilizes channel tokens to effectively reconstruct image patches. Experiments on satellite and microscopy datasets, CHAMMI, JUMP-CP, and So2Sat, show that ChA-MAEViT significantly outperforms state-of-the-art MCI-ViTs by 3.0-21.5\%, highlighting the importance of cross-channel interactions in MCI. Our code is publicly available at 
\textcolor{pinkcolor2}{\href{https://github.com/chaudatascience/cha_mae_vit}{\url{https://github.com/chaudatascience/cha_mae_vit}}}.
\end{abstract}

%% file: 2_introduction.tex
\section{Introduction}
\label{sec:intro}

\begin{figure}
  \centering
\includegraphics[width=.95\linewidth]{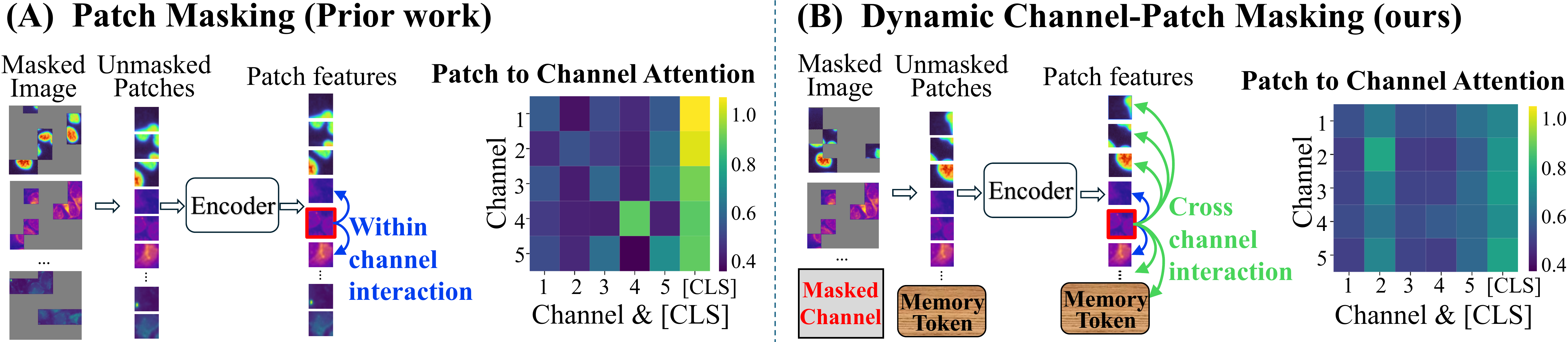}
\vspace{-2mm}
  \caption{\textbf{Image patch interactions in MCI.} \textbf{(A) Prior Work on MCI-MAEs} (\eg, \camae~\cite{kraus2024masked}) employ random patch masking to train the model to reconstruct the masked patches. In the attention map, where each row represents the average attention score of all patches in a channel towards other channels and the \methodfont{[CLS]} token (last column), we show each patch primarily attends to its own channel (the diagonal) and the \methodfont{[CLS]} token.  This suggests that patch masking may not effectively promote cross-channel interactions in MCI. \textbf{(B) In contrast, \ourmask{} (ours)} encourages more interactions between patches across different channels by using both channel and patch masking. Also, \textit{memory tokens} serve as long-term memory to help information aggregation across channels. Its attention pattern demonstrates a more uniform distribution across channels, indicating that each image patch can learn more meaningful interactions.}
\label{fig:motivation}
   \vspace{-4mm}
\end{figure}

Visual encoders generally process fixed-channel inputs, such as RGB, during both training and testing phases~\cite{inception,resnet2,efficientnet,mobilenet,swin,liu2022convnet,hiera,liu2025vmamba}. However, in satellite imaging~\cite{mediatum1483140}, robotic sensing~\cite{zou2024unim}, cell microscopy~\cite{chen2023chammi,chadavit}, and medical imaging~\cite{chen2017cross}, the input data can vary in both the number and type of channels. These varying channel configurations arise from differences in sensor modalities, acquisition settings, or experimental conditions. Multi-Channel Imaging (MCI) models are designed to learn feature representations from heterogeneous channels whose type and number vary during both training and testing~\cite{chen2023chammi,chadavit}. This adaptability allows a single model to effectively support diverse channel configurations, thereby reducing computational costs and minimizing the risk of overfitting~\cite{chen2023chammi}.

Prior MCI-Masked Autoencoder (MAE) models (\eg, \cite{kraus2024masked,kenyon-dean2025vitally}) demonstrated promise in capturing local spatial structures by learning to reconstruct randomly masked patches in multi-channel images.
These approaches assume that images exhibit significant redundancy across channels, enabling the reconstruction of masked patches using unmasked patches from similar channels. While this assumption works well for natural images, where color channels typically show strong correlations, it presents challenges in MCI scenarios. In these situations, complementary sensors (\eg, multispectral and LiDAR) may capture distinct physical properties with minimal feature overlap. As shown in \cref{fig:motivation}(a), when using patch masking, the patch-to-channel attention concentrates on its own channel (the diagonal) and the \methodfont{[CLS]} token. This indicates that prior MAE methods mainly focus on visible patches within the same channel, neglecting cross-channel feature interactions, restricting the model's ability to learn useful features that require information from multiple channels.

\begin{figure}
  \centering
\includegraphics[width=.88\linewidth]{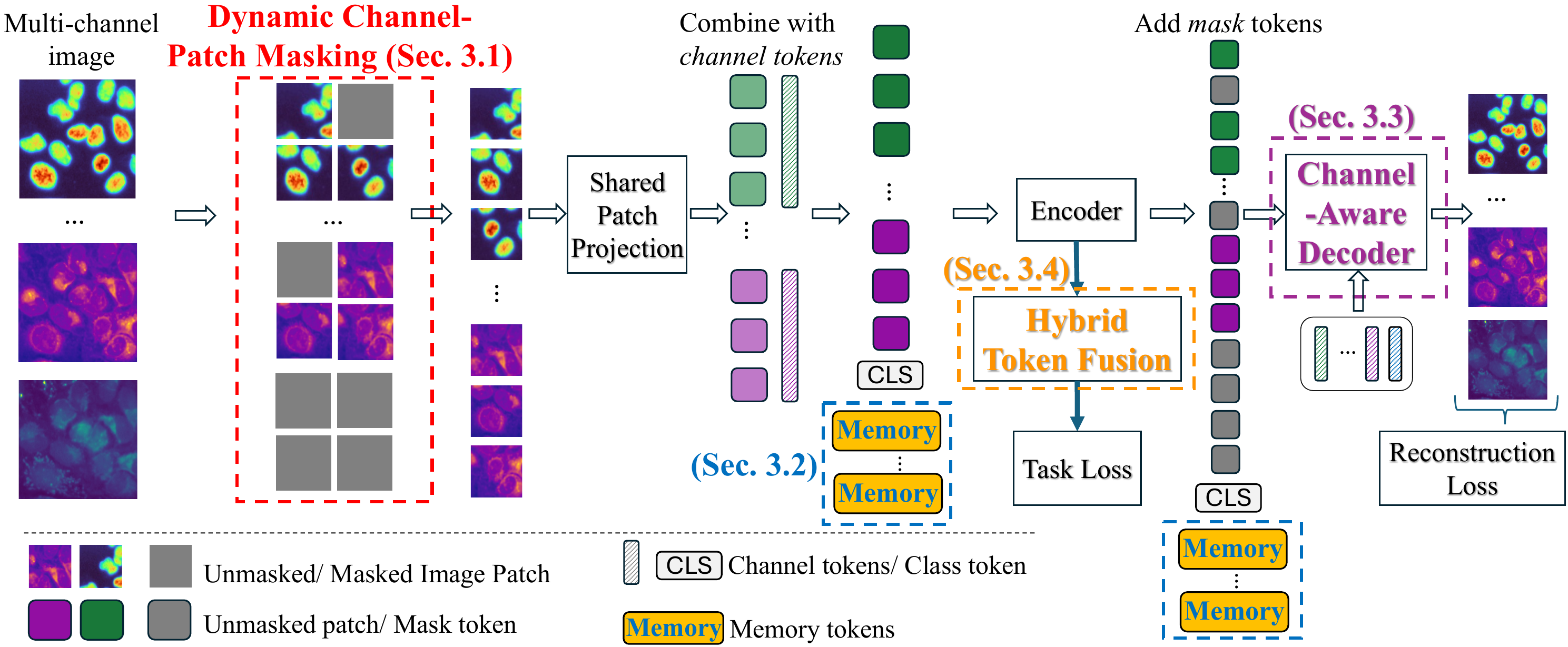}
\vspace{-2mm}
  \caption{Our \textbf{\ours} approach enhances cross-channel learning via four key components: 1) \textbf{\textcolor{redcolor}{\ourmask}}, which compels the model to reconstruct varying proportions of missing channels and patches, thus improving interactions across channels and robustness to the absence of some channels (\cref{sec:mask_strategy}). 2) \textbf{\textcolor{bluecolor}{Memory Tokens}}, which act as long-term memory to facilitate information sharing between channels (\cref{sec:memory_tokens}). 3) To reconstruct the masked patches and channels, we use a \textbf{\textcolor{purplecolor}{Channel-Aware Decoder}} that leverages channel tokens for image reconstruction, enhancing performance while minimizing computational costs (\cref{sec:channel_aware_decoder}). 4) A \textbf{\textcolor{orangecolor}{Hybrid Token Fusion}} module, which combines fine-grained patch tokens with a global \methodfont{[CLS]} token to improve feature representation (\cref{sec:hybrid_token_fusion}).}
\label{fig:overview_architecture}
   \vspace{-4mm}
\end{figure}

To address this challenge, we introduce \oursfullnamewithabbre, 
which enhances cross-channel learning in MCI through four key improvements summarized in \cref{fig:overview_architecture}. 
First, we propose \ourmaskwithfontandabbr, which adaptively masks both patches and channels during training, compelling the model to reconstruct these missing patches and channels using the remaining patches (\cref{fig:overview_architecture}, left). 
\ourmaskabbr{} adjusts the channel masking ratio to enhance feature learning and robustness to missing channels during inference. 
\cref{fig:motivation}(b) shows our approach evenly redistributes patch attention scores across channels, highlighting improved cross-channel interactions. However, in MCI, reconstructing masked channels is challenging since each channel may encode unique features that are not easily inferred from others.

To enable the model to retain global information across all channels during reconstruction, we introduce the use of \textit{memory tokens} in MCI (\cref{fig:overview_architecture}, middle). Inspired by register tokens~\cite{darcet2024vision}, which are extra tokens added in the input to reduce artifacts in the feature maps of ViTs, these learnable embeddings serve as long-term memory to retain key cross-channel information in both the encoder and decoder. 
In addition, we use a hybrid token fusion module in the encoder to combine fine-grained patch tokens with the global class token for a richer representation (\cref{fig:overview_architecture}, middle). 

Finally, we employ \methodfont{Channel-Aware Decoder} that simultaneously processes patch tokens from all channels. Existing methods often rely on separate decoders for each channel~\cite{bachmann2022multimae, zhang2024multimodal, kraus2024masked}, which do not scale well with many input channels. Our shared lightweight decoder utilizes channel tokens to provide channel-specific behavior and memory tokens to improve cross-channel feature reconstruction, enhancing performance while reducing computational costs.

Most of our experiments use both self-supervised (SSL) and supervised learning to boost performance, showing a similar complementary benefit from combining them (\eg,~\cite{liang2022supmae,cheng2021self,bai2022joint,xin2024masked}). However, \cref{sec:additional_results_appendix} also evaluates SSL by itself, where we show our approach still outperforms prior work. 

The most relevant work to ours focuses on enhancing cross-channel interactions~\cite{gao2020channel,zhou2022understanding,hu2018squeeze,yang2019cross}. However, these approaches mainly use single-modality images, such as RGB, where channel similarities facilitate strong correlations. As noted earlier, this often does not generalize to the more complex MCI domain where channels convey varying information, \eg, a robot sensor can contain LiDAR, RGB, and thermal camera. Thus, MCI requires a balance between preserving unique information in each channel and effectively modeling the complex relationships between channels. MCI models must also be robust to missing channels to support varying channel configurations~\cite{chen2023chammi,chadavit}. This means that cross-channel interaction should not only improve the overall representation but also involve learning useful redundancy across channels.
\ours{} addresses these challenges by introducing \ourmaskabbr{} to enhance cross-channel interaction while allowing the model to learn important channel-specific features, resulting in a more robust and higher-performing model. 

Our key contributions are as follows:
\vspace{-2mm}
\begin{itemize}[nosep,leftmargin=*]
\item We explore \ours, which integrates MAEs into MCI-ViTs for improved robustness that reports a gain of \ourimprove{} across three diverse MCI datasets, CHAMMI~\cite{chen2023chammi}, JUMP-CP~\cite{jumpcpSrinivas}, So2Sat~\cite{mediatum1483140}, and consistent gains on Cloud-38~\cite{mohajerani2019cloud}.
\item   We propose \ourmaskwithfont{}, which adaptively masks both channels and patches during training, requiring the model to reconstruct them 
to boost cross-channel  interaction.

\item We introduce a \methodfont{Channel-Aware Decoder} that uses channel tokens to reconstruct image patches more efficiently, thereby enhancing performance for downstream tasks.

 \item We use \textit{memory tokens} $-$ learnable vectors to retain cross-channel information and aggregate channel-specific features, and \textit{hybrid token fusion module} to merge the \methodfont{[CLS]} token with fine-grained patch tokens for enhanced feature representation.

\end{itemize}

%% file: 3_related_work.tex
\section{Related Work}

Many MCI methods do not use self-supervised objectives~\cite{phamDiChaVit2024,bao2024channel,chadavit}, but rather focus on aspects like diversity and robustness to missing channels.  However, as noted in the introduction, self-supervised objectives like Masked Autoencoders (MAEs) can provide complementary information to further boost performance. MAEs themselves are often built for RGB images (\eg,~\cite{he2022masked,liang2022supmae,xie2022simmim}) or those that are also strongly aligned like thermal images~\cite{zhang2024multimodal}, based on assumptions of strong correlations between channels does not extend to many MCI images. Many methods in other domains followed suit, performing channel-wise masking on protein markers~\cite{sims2023mim} or masked both of spatial and spectral channels~\cite{lin2023ss,wang2024hsimae}. However, these methods rely on a fixed ratio of masked channels, which can be inadequate for the varying channel configurations found in MCI. In addition, prior work uses dual-branch architectures that treats spatial and channel representations independently, limiting meaningful interactions between channels, especially when spatial alignment is not guaranteed. In contrast, we dynamically mask a variable number of channels each time. By combining both channel and patch-level masking within a single encoder, we achieve better feature interaction.

The work most similar to ours is \camae{}~\cite{kraus2024masked}, which randomly masks a fixed proportion of patches from all channels and employs distinct decoders for reconstruction. In contrast, \ours{} employs both channel and patch-level masking with a shared decoder, and memory tokens serve as long-term memory to facilitate feature interactions. 

There are other types of self-supervised learning methods beyond MAEs that have been explored in other settings (\eg,~\cite{geiping2023cookbook,chen2020simple,chen2020big,dwibedi2021little,chen2021exploring,grill2020bootstrap,caron2021emerging,he2022masked,xie2022simmim,zhou2021ibot,oquab2024dinov,assran2022masked}).  However, similar to MAEs, they often make assumptions that do not generalize to MCI images.  For example, self-distillation methods (\eg, SimSiam~\cite{chen2021exploring}, \methodfont{BYOL}~\cite{grill2020bootstrap}, \dino~\cite{caron2021emerging}) process two different views through two encoders, and then map one view to the other using a predictor network.
These methods often rely on complex augmentations designed for natural images to generate dual views, such as color jittering and Gaussian blur. These are less effective for MCI, where heterogeneous channels (\eg, LiDAR, thermal, RGB) carry distinct physical properties. Such augmentations can distort critical features like depth or intensity, hindering the model's ability to capture cross-channel dependencies essential for MCI~\cite{roddan2024calibration,patnala2023generating}. This makes masked modeling techniques attractive (\eg, MAE~\cite{he2022masked}, SimMim~\cite{xie2022simmim}), as they do not rely on augmentations that may be challenging to apply with more complex and varied images.

%% file: 4_method.tex
\section{Channel-Aware MAE for Improved Cross-Channel Learning}
\label{sec:method}

Given a multi-channel image (MCI) denoted as $X$, which consists of various channels $c_i \in C$, our objective is to train a model $M$ that takes $X$ as input to generate representations and/or predictions. Following \cite{chadavit, bao2024channel, phamDiChaVit2024, chen2023chammi}, we focus on the MCI setting where the model $M$ is trained using all available channels $C$ but tested with a subset of those channels $C_{\mathrm{test}} \subseteq C$. 
We describe the details of the four main components of our proposed framework, which is illustrated in \cref{fig:overview_architecture}.

\subsection{\ourmask}
\label{sec:mask_strategy}

As shown in \cref{fig:motivation}(a), image patches in prior work seem to attend mostly to their own channels while neglecting the others, potentially missing rich interactions among channels. To mitigate this issue, we propose \ourmaskwithfontandabbr, which integrates both patch and channel-level masking strategies. We start creating image patches (\cref{sec:patching_details}), which results in $n$ tokens per channel, each with $d$-dimensional embeddings, for $c$ channels.

Our masking strategy consists of two components: \textit{random patch masking} and \textit{dynamic channel masking}. First, \textit{random patch masking}, aims to generate a mask \(\mathbf{p\_mask} \in \{0, 1\}^{n \times c}\) that applies a fixed masking ratio \(r_p\) (\eg, $75$\%) to mask patches independently across each channel. Specifically, for each channel $j$, we uniformly sample a set of \(\lfloor n \cdot r_p \rfloor\) positions from the \(n\) available patch positions, denoted \(S_j \subset \{1, 2, \dots, n\}\) with \(|S_j| = \lfloor n \cdot r_p \rfloor\). We then mask these patches as
$\mathbf{p\_mask}_{i,j} =
\begin{cases} 
1 & \text{if } i \in S_j \\
0 & \text{otherwise}
\end{cases}
$, where \(\mathbf{p\_mask}_{i,j} = 1\) indicates the patch at position \(i\) of channel \(j\) is masked, and \(\mathbf{p\_mask}_{i,j} = 0\) indicates unmasked.
All channels have the same number of masked patches, but locations vary per channel because they are independently sampled, as opposed to prior work that first generates the mask for one channel and then replicates it to all other channels~\cite{nedungadi2024mmearth,wang2024hsimae}. 

The second component, \textit{dynamic channel masking}, aims to generate a mask \(\mathbf{c\_mask} \in \{0, 1\}^{n \times c}\) that adaptively mask out some channels. 
Here, the term "dynamic" refers to the varying number of channels masked during training, rather than adaptation based on the specific input data.
We start by uniformly sampling the number of channels to be masked, denoted as $k \sim \mathcal{U}\{0, 1, \ldots, c-1\}$.
Then, we uniformly sample a set of $k$ channels, denoted as 
$\mathcal{C}' \subset \{1, \dots, c\}$ with $|\mathcal{C}'|=k$, and mask these channels out as 
$\mathbf{c\_mask}_{\star,   j} = \begin{cases}
\mathbf{1}^n & \text{if } j \in \mathcal{C}' \\
\mathbf{0}^n & \text{otherwise}
\end{cases}
$. This masking approach is inspired by \methodfont{Hierarchical Channel Sampling (HCS)}~\cite{bao2024channel}. However, unlike \methodfont{HCS}, which serves as a channel dropout technique that completely removes selected channels, our approach employs masked channels as supervisory signals, designating them as labels for the reconstruction process. This enables the model to directly learn inter-channel relationships, thus enhancing greater cross-channel feature interaction.

Finally, we integrate these two components for training images in our \ourmaskabbr{} strategy by introducing hyperparameters \(\mathrm{p}_\mathrm{patch}\), \(\mathrm{p}_\mathrm{channel}\) $\in (0,1)$. These values divide the unit interval into three sections to randomly determine how to use one or the other type of mask, as follows: 
\begin{equation}
\mathbf{mask} = \begin{cases}
\mathbf{p\_mask} & \text{if } s < \mathrm{p}_\mathrm{patch} \\
\mathbf{c\_mask} & \text{if } \mathrm{p}_\mathrm{patch} \leq s < \mathrm{p}_\mathrm{patch} + \mathrm{p}_\mathrm{channel} \\
\mathbf{p\_mask} \lor \mathbf{c\_mask} & \text{otherwise}
\end{cases}
\end{equation}

where $s$ is a selection value chosen uniformly at random. In practice, we found that optimizing the model with these two masking strategies simultaneously can be difficult, as it may lead to excessive information loss, making image reconstruction challenging. To address this, we adjust the hyperparameters to alternate between the two masking strategies separately. Specifically, setting \(\mathrm{p}_\mathrm{patch} = \mathrm{p}_\mathrm{channel} = 0\)  merges both patch and channel masks into a unified mask, and setting \(\mathrm{p}_\mathrm{patch} = \mathrm{p}_\mathrm{channel} = 0.5\)  allows the model to switch between patch and channel masks. We adopt these two straightforward configurations for all our experiments. Refer to \cref{sec:mask_algorithm} for the procedure of \ourmaskabbr{}, and \cref{sec:dcp_masking_more_analysis} for more analysis on the hyper-parameters.

\subsection{Memory Tokens}
\label{sec:memory_tokens}
 
Reconstructing masked channels is challenging due to their inherent differences, since each channel encodes unique features not easily inferred from others, \eg, reconstructing LIDAR from RGB.
Inspired by the concept of register tokens \cite{darcet2024vision}, which are used to reduce artifacts in the feature maps of ViTs, we introduce \textit{memory tokens} for MCI (\cref{fig:overview_architecture}, middle). These memory tokens are learnable embeddings that serve as long-term memory, allowing for the storage of global information across all channels. During training, these tokens gather channel-specific features using self-attention mechanisms, helping the model retain and propagate information across layers that might otherwise be lost due to masking. Additionally, these tokens assist in decoding image patches to facilitate the reconstruction process. During inference, memory tokens enable the model to retrieve stored context, effectively addressing the issue of missing channels. Similar to the \methodfont{[CLS]} token, memory tokens are incorporated into the input during both training and inference. However, while the \methodfont{[CLS]} token is utilized as the final representation, the memory tokens are excluded. 

Formally, we prepend $l$ memory tokens $\{\mathbf{M_i}\}_{i \in [l]}$ into the input sequence, resulting in
$
[\mathbf{t}_{\mathrm{CLS}}; \mathbf{M}_1; \ldots; \mathbf{M}_l; \mathbf{t}_{1,1}; \mathbf{t}_{2,1}; \ldots; \mathbf{t}_{n,c}]$, where $\mathbf{t}_{\mathrm{CLS}}$ and $\mathbf{t}_{i,j}$ are the class token and patch token at $i$-th location of $j$-th channel, respectively. 
After masking some patches (\cref{sec:mask_strategy}), the remaining patches are fed into a transformer encoder. 
Following \cite{bao2024channel,phamDiChaVit2024}, spatial information is incorporated through learnable positional embeddings, while channel-specific properties are captured by special \textit{channel tokens}, each represented as a learnable embedding. These channel tokens are concatenated with the patch tokens and jointly processed by the transformer encoder and decoder.

\subsection{Channel-Aware Decoder}
\label{sec:channel_aware_decoder}
After processing the unmasked tokens with the encoder, we reconstruct the masked patches using the unmasked ones with a \methodfont{Channel-Aware Decoder}. Unlike prior MCI-MAE methods that use separate decoders for each channel or modality~\cite{bachmann2022multimae, zhang2024multimodal, kraus2024masked}, we employ a single decoder to process tokens from all channels simultaneously (\cref{fig:overview_architecture}, right). This scales better to the number of input channels (up to $18$ in our experiments), while also boosting performance.

Let the output sequence from the encoder be
$[\mathbf{\hat{t}}_{\mathrm{CLS}}; \mathbf{\hat{M}}_1; \ldots; \mathbf{\hat{M}}_l; \mathbf{\hat{t}}_1; \mathbf{\hat{t}}_2; \ldots; \mathbf{\hat{t}}_{v}]$,
where $v$ denotes the number of visible (\ie, unmasked) patches fed into the encoder. This sequence is shorter than the original image patch length due to the missing masked patches. To reconstruct these masked patches, we utilize $u = (n \cdot c - v)$ mask tokens $\mathbf{m}_i$, where each mask token is a shared, learned vector representing the missing patch.

\smallbreak
\noindent \textbf{Incorporating Channel-Specific Information.}
To reconstruct channel-specific information, we combine patch tokens, whether visible ($\mathbf{\hat{t}}_i$) or masked ($\mathbf{m}_i$), with their corresponding \textit{channel tokens}, which are also optimized by the encoder. We define $f(\cdot)$ as a function that returns the corresponding \textit{channel token} for a given patch token. The input to our Channel-Aware Decoder is thus:

$\mathbf{\hat{T}} = [\mathbf{\hat{t}}_{\mathrm{CLS}}; \mathbf{\hat{M}}_1; \ldots; \mathbf{\hat{M}}_l; \mathbf{\hat{t}}_1 + f(\mathbf{\hat{t}}_1); \mathbf{\hat{t}}_2 +f(\mathbf{\hat{t}}_2) ; \ldots; \mathbf{\hat{t}}_{v} +f(\mathbf{\hat{t}}_v); \mathbf{m}_1 +f(\mathbf{m}_1), \dots, \mathbf{m}_u + f(\mathbf{m}_u)]$

Incorporating channel tokens provides a more channel-specific context, which enhances the reconstruction process.
Additionally, we incorporate learnable positional embeddings, which are shared with the encoder, to provide information about the position of each patch in the image.
\smallbreak
\noindent \textbf{Patch Reconstruction.}
We pass the token sequence $\mathbf{\hat{T}}$ into several Transformer Blocks, followed by a Decoder Head to reconstruct the pixels of each image patch. Let \(\mathbf{T} \in \mathbb{R}^{(n \cdot c) \times d}\) represent the output of the patch tokens from the final Transformer Block. The Decoder Head is a linear layer \(\mathbf{W_\mathrm{decoder}} \in \mathbb{R}^{d \times p^2}\), where \(p\) is the image patch size. The pixel reconstruction of the $i$-th patch of the $j$-th channel, is computed as  $\mathbf{\hat{x}}_{i,j} = (\mathbf{T}_{i,j} \cdot \mathbf{W_\mathrm{decoder}}) \in \mathbb{R}^{p^2}$.
We found that a lightweight decoder with just $1-2$ Transformer Blocks is sufficient, aligning with findings from prior work~\cite{he2022masked, liang2022supmae}.

\smallbreak
\noindent \textbf{Reconstruction Loss.}
We use standard $L2$ loss on the image pixel, and the Fourier space $L1$ loss introduced in \cite{kraus2024masked}, both computed only on the masked patches.
Let $P= \sum_{i=1}^{n}  \sum_{j=1}^{c} \mathbf{mask}_{i,j}$ be the number of masked patches of the input image. Given $\mathbf{x}_{i,j}, \mathbf{\hat{x}}_{i,j} \in \mathbb{R}^{p^2}$, the original pixels of patch $i$ in channel $j$, and its reconstruction, respectively, the reconstruction loss is as follows:
\resizebox{\columnwidth}{!}{$\mathcal{L}_{\mathrm{pixel}} = \frac{1}{P} \sum_{i=1}^{n}  \sum_{j=1}^{c} \mathbf{mask}_{i,j} \cdot L_2(\mathbf{x}_{i,j}, \mathbf{\hat{x}}_{i,j});\quad \mathcal{L}_{\mathrm{{fourier}}} = \frac{1}{P} \sum_{i=1}^{n}  \sum_{j=1}^{c} \mathbf{mask}_{i,j}\cdot L_1(|\mathcal{F}(\mathbf{x}_{i,j})|, |\mathcal{F}( \mathbf{\hat{x}}_{i,j})|)$}
\vspace{-2mm}
\begin{equation}
\mathcal{L}_{\mathrm {recon}} = (1 - \lambda_f) \cdot \mathcal{L}_{\mathrm{pixel}} + \lambda_f \cdot \mathcal{L}_{\mathrm{fourier}},
\label{eq:reconloss}
\end{equation}
where $|\mathcal{F}|$ is the amplitudes of Fast Fourier Transform. Following~\cite{kraus2024masked}, we set a fixed $\lambda_f$ to $0.01$.

\subsection{Hybrid Token Fusion}
\label{sec:hybrid_token_fusion}
Prior work either utilizes the \methodfont{[CLS]} token~\cite{darcet2024vision,oquab2023dinov2} or average pooling of patch tokens~\cite{liang2022supmae,kraus2024masked} for downstream tasks. We found that incorporating both in our framework yields better performance (\cref{fig:overview_architecture}, middle). Given the output sequence from the encoder, we use a learnable query vector $\mathbf{q}_{\text{patch}}$ to interact with the patch tokens. The query attends to the patch tokens through cross-attention, and then combines with the \methodfont{[CLS]} token to generate a fused representation $f_{\mathrm{fusion}} = \hat{t}_{\mathrm{CLS}} \odot \sigma(\mathrm{CrossAttention}(\mathbf{q}_{\mathrm{patch}}, \mathbf{T}_p))$,
where $\mathbf{T}_p=[\mathbf{\hat{t}}_1; \mathbf{\hat{t}}_2; \ldots; \mathbf{\hat{t}}_{v}]$, $\odot$ represents the elementwise product, and $\sigma$ the sigmoid function. The fused representation, $f_{\mathrm{fusion}}$, integrates global context from the \methodfont{[CLS]} token while also being refined by detailed spatial information from the patch tokens. We then use a multi-layer perceptron ($\mathrm{MLP}$) with GELU activation~\cite{hendrycks2016gaussian} to enhance the fusion representation and tailor it for downstream tasks:
$f_{\mathrm{final}} = \mathrm{Linear}(\mathrm{GELU}(\mathrm{Linear}(f_{\mathrm{fusion}}))).
\label{eq:fusion}$

\noindent\textbf{Training Objective.}
The output of \cref{eq:fusion} is then used to compute the task loss, \eg, cross-entropy for classification.
Our final loss consists of three components: the primary loss for the specific task $\mathcal{L}_{\mathrm{task}}$, regularization term $\mathcal{L}_{\mathrm {d}}$, and reconstruction loss $\mathcal{L}_{\mathrm {recon}}$ (\cref{eq:reconloss}) as follows:
\begin{equation}
\mathcal{L}_{\mathrm{final}} = (1-\lambda_{\mathrm{recon}} ) \cdot (  \mathcal{L}_{\mathrm{task}} + \lambda_{\mathrm{d}} \cdot  \mathcal{L}_{\mathrm {d}} ) +  \lambda_{\mathrm{recon}} \cdot  \mathcal{L}_{\mathrm {recon}},
\label{eq:finalloss}
\end{equation}
where $\lambda_{\mathrm{d}}$ is to balance the task loss and the regularization term, and $\lambda_{\mathrm{recon}}$ is to balance the reconstruction loss with other losses. For regularization $\mathcal{L}_{\mathrm {d}}$, we use the losses introduced in \cite{phamDiChaVit2024}, in which \methodfont{Channel Diversification Loss} applies to the channel tokens, and \methodfont{Token Diversification Loss} applies to the patch tokens to encourage diversity learning. 
We set a small $\lambda_{\mathrm{d}}$ (\eg, $0.001$) as suggested by \cite{phamDiChaVit2024}, and a fixed $\lambda_{\mathrm{recon}}=0.99$ for all experiments.

%% file: 5_experiments.tex
\section{Experiments}
\noindent\textbf{Datasets.} We evaluate on three cell microscopy and satellite datasets. \textit{(i)} \textbf{CHAMMI}~\cite{chen2023chammi}, a channel-adaptive benchmark consists of varying-channel images sourced from WTC-11, HPA and CP, with $3$, $4$, and $5$ channels respectively. Together, these three datasets contain $220$K microscopy images, in which, $100$K images are for training, while the rest for testing across domain generalization tasks. \textit{(ii)} \textbf{JUMP-CP}~\cite{jumpcpSrinivas} is a cellular imaging dataset, where each image has $8$ channels, consisting of $5$ fluorescence and $3$ brightfield channels. We use compound perturbation plate BR00116991, which contains $127$K training, $45$K validation, and $45$K test images, across $161$ classes. \textit{(iii)} \textbf{So2Sat}~\cite{mediatum1483140} contains synthetic aperture radar and multispectral optical image patches from remote sensing satellites, with $18$ channels ($8$ from Sentinel-1, $10$ from Sentinel-2) and $17$ climate zone classes. We use the city-split version, consisting of $352$K training, $24$K validation, and $24$K test images.

\input{tables/main_result_sup}

\noindent{\bf Baseline methods.} 
We compare to the following methods:
 \vspace{-2mm}
\begin{itemize}[nosep,leftmargin=*]
    \item \textbf{\depthwisevit{}}~\cite{chen2023chammi} processes each input channel through a depthwise convolution layer, averages the filtered features, and feeds them into a \vit{} backbone.

\item \textbf{\templatevit{}}~\cite{plummerNPAS2022,savarese2018learning,phamMixtureGrowth2024} learns channel weights via shared parameter templates, forming a patch projection layer for a \vit{} backbone.

\item \textbf{\hypervit{}}~\cite{haHypernetworks2016} uses a neural network to generate channel-specific weights, concatenating them into a patch projection layer for a \vit{} backbone.

\item \textbf{\camae}~\cite{kraus2024masked} extends MAE for multi-channel imaging. We add task loss to adapt this baseline.

\item \textbf{\chadavit{}}~\cite{chadavit} employs a shared projection for channel-wise feature extraction, combining the tokens with positional and channel embeddings for \vit{} processing.

\item \textbf{\channelvit{}}~\cite{bao2024channel} Similar to \chadavit{}, while incorporating Hierarchical Channel Sampling.

\item \textbf{\dichavit{}}~\cite{phamDiChaVit2024} improves \channelvit{} with regularization terms and diverse channel sampling.
\end{itemize}
\noindent Additionally, we investigate the effect of combining the best supervised baseline, \dichavit{}, with the following SSL to evaluate the importance of incorporating self-supervision:  \methodfont{MAE}~\cite{he2022masked,kraus2024masked},
 \simclr~\cite{chen2020simple,chen2020big}, \simsiam~\cite{chen2021exploring}, \ibot~\cite{zhou2021ibot}, and \dinotwo~\cite{oquab2024dinov}. More details in \cref{sec:ssl_method_details}. 

\noindent\textbf{Implementation details.} 
All baselines employ ViT-S ($21$M) as the backbone. In our method, we remove one transformer block from the encoder to accommodate the decoder, ensuring that our approach maintains the same number of parameters as the baselines. Refer to \cref{sec:implementation_details_appendix} for details.

\noindent\textbf{Metrics.} We reported top-1 classification accuracy for the classification tasks on So2Sat~\cite{mediatum1483140} and JUMP-CP~\cite{jumpcpSrinivas}. For the representation learning tasks on CHAMMI~\cite{chen2023chammi}, following \cite{phamDiChaVit2024}, we report the average F1 scores across WTC-11 and HPA.

\subsection{Results}

\cref{tab:main_result_sup} compares \ours{} with different MCI-ViTs. \ours{} significantly outperforms the baseline models by an average of $10.0\%$ across three datasets. Specifically, \ours{} exceeds the state-of-the-art model \dichavit{} by $5.0\%$ on CHAMMI, a channel-adaptive benchmark. For JUMP-CP and So2Sat, we conducted evaluations in both \textit{Full} (using all channels) and \textit{Partial} (using subsets of channels) settings. In the full setting, our model surpasses the next best models by $21.5\%$ and $3.0\%$, respectively. 
Additionally, \ours{} demonstrates its robustness in the partial settings, where we test JUMP-CP using five fluorescence channels and So2Sat using eight Sentinel-1 channels, showing improvement of $10.0\%$ and $4.5\%$ points respectively over prior work.


\cref{tab:main_result_ssl} presents the impact of integrating various SSL methods with the top-performing supervised approach, \dichavit~\cite{phamDiChaVit2024}. In the last row, we train \dichavit{} to reconstruct the masked patches using only our \ourmaskabbr{}, without incorporating \textit{memory tokens}, \methodfont{Hybrid Token Fusion}, or \methodfont{Channel-Aware Decoder}. While the combination with \methodfont{MAE} gives the highest performance, \methodfont{DCP} still significantly outperforms the best SSL-enhanced variant by $0.6-5.6\%$ across three datasets.
The performance gap highlights the effectiveness of \methodfont{DCP}, which substantially exceeds the improvements gained from solely combining with SSL objectives.
Additionally, by only processing the unmasked patches, our method achieves significantly faster runtime compared to other SSL methods, \eg, $6X$ faster than \dinotwo{}, making it both performant and computationally efficient for MCI tasks. Refer to \cref {sec:ssl_method_details} and Appendix \cref{fig:flops} for more discussion on runtime and  FLOPs comparisons.

\input{tables/main_result_ssl}

\input{tables/segmentation}

\input{tables/ours_ablation}

To further demonstrate our approach's ability to generalize, \cref{tab:segmentation} reports segmentation performance on 38-Cloud~\cite{mohajerani2019cloud}, complementing our results on the classification (So2Sat~\cite{mediatum1483140}, JUMP-CP~\cite{jumpcpSrinivas}) and representation learning tasks (CHAMMI~\cite{chen2023chammi}) reported earlier.  \ours{} outperforms all baselines across evaluation metrics, demonstrating its ability to boost multi-channel learning.

\noindent \textbf{Ablation study of \ours.}  
\cref{tab:main_ablation} presents the
model's performance when a component is replaced. Specifically, "w/o DCP Masking" indicates that we replace our DCP Masking with random patch masking in \methodfont{CA-MAE}~\cite{kraus2024masked}, "w/o Hybrid Token Fusion" uses \methodfont{[CLS]} token instead of Hybrid Token Fusion module, "w/o Memory Tokens" means no memory token is being used, and "w/o Channel-Aware Decoder" indicates replacing our Channel-Aware Decoder with \camae's Separate Decoders. The results highlight the critical role of each of the components, especially DCP Masking, as its removal has the most detrimental effect on performance.

%% file: tables/main_result_sup.tex
\begin{table*}[t]
  \centering
  \setlength\tabcolsep{2.8pt}
  \caption{\textbf{Comparison of channel adaptive models}. We report the mean accuracy with standard deviation on the test set of three runs. "Full" refers to inference on all channels, while "Partial" means inference on a subset of channels. \ours{} outperforms other baselines consistently across three datasets. Note: all models contain the same number of parameters, except \camae{} due to its separate decoders (\eg, $4X$ parameters on So2Sat).}
  \vspace{-2mm}
  \begin{tabular}{rc*{5}{c}}
    \toprule
     &  CHAMMI~\cite{chen2023chammi} & \multicolumn{2}{c}{JUMP-CP~\cite{jumpcpSrinivas}} & \multicolumn{2}{c}{So2Sat~\cite{mediatum1483140}} \\
     \cmidrule(r){2-2} \cmidrule(r){3-4} \cmidrule(r){5-6} 
     Model     & Avg score  & Full   & Partial &  Full   & Partial \\
    \midrule
    
   \hypervit~\cite{haHypernetworks2016}  & 56.08\footnotesize{${\pm0.41}$} & 47.07\footnotesize{${\pm0.47}$} & 42.43\footnotesize{${\pm0.65}$} & 60.73\footnotesize{${\pm0.24}$} & 41.88\footnotesize{${\pm0.85}$}\\
   
    \depthwisevit~\cite{chen2023chammi}  & 61.80\footnotesize{${\pm0.43}$} & 49.86\footnotesize{${\pm0.45}$} & 44.98\footnotesize{${\pm0.71}$} & 60.41\footnotesize{${\pm0.22}$}  & 43.41\footnotesize{${\pm1.10}$} \\
    \templatevit~\cite{plummerNPAS2022,savarese2018learning,phamMixtureGrowth2024}  & 58.16\footnotesize{${\pm0.42}$} & 52.48\footnotesize{${\pm0.27}$} & 43.85\footnotesize{${\pm0.73}$} & 55.86\footnotesize{${\pm0.10}$} & 37.28\footnotesize{${\pm0.34}$} \\
    
    \camae~\cite{kraus2024masked}  \methodfont{+ Sup. loss} & 59.15\footnotesize{${\pm0.28}$} & 69.54\footnotesize{${\pm0.12}$}  & 20.93\footnotesize{${\pm0.25}$} & 64.21\footnotesize{${\pm0.41}$} & 15.75\footnotesize{${\pm0.83}$}\\

   \chadavit~\cite{chadavit} &  63.93\footnotesize{${\pm0.42}$} & 65.03\footnotesize{${\pm0.98}$} & 42.15\footnotesize{${\pm2.33}$} & 56.98\footnotesize{${\pm0.46}$} & 12.38\footnotesize{${\pm2.03}$}\\
    
    \channelvit~\cite{bao2024channel} &   
 64.97\footnotesize{${\pm0.58}$} & 67.51\footnotesize{${\pm0.35}$} & 56.49\footnotesize{${\pm0.53}$} & 61.03\footnotesize{${\pm0.17}$} & 46.16\footnotesize{${\pm0.40}$}\\
      
   \dichavit~\cite{phamDiChaVit2024}  & 69.77\footnotesize{${\pm0.44}$} & 69.19\footnotesize{${\pm0.47}$} & 57.98\footnotesize{${\pm0.41}$} & 63.36\footnotesize{${\pm0.11}$} & 47.76\footnotesize{${\pm0.23}$} \\

\midrule
  
     \textbf{\oursbr}  & \textbf{74.63}\footnotesize{${\pm0.54}$} & \textbf{90.73}\footnotesize{${\pm0.14}$} & \textbf{68.05}\footnotesize{${\pm0.21}$} & \textbf{67.44}\footnotesize{${\pm0.38}$} & \textbf{52.11}\footnotesize{${\pm0.49}$} \\

    \bottomrule
  \end{tabular}
    \label{tab:main_result_sup}
\end{table*}

%% file: tables/main_result_ssl.tex
    
    



\begin{table}[t]
  \centering
  \caption{\textbf{SSL methods when combined with the best supervised baseline \dichavit~\cite{phamDiChaVit2024}}. The last row shows the combination of \dichavit{} with \ourmaskwithfont. Incorporating SSL in \dichavit{} results in improvements. Best result with our masking strategy. \textbf{Boldfaces} and \underline{underlines} indicates best and second best numbers, respectively.}
  \label{tab:main_result_ssl}
  \begin{tabular}{r c *{5}{c} @{}}
    \toprule
     & \multirow{2}{*}{CHAMMI} & \multicolumn{2}{c}{JUMP-CP} & \multicolumn{2}{c}{So2Sat} \\
     \cmidrule(r){3-4} \cmidrule(r){5-6} 
      \dichavit~\cite{phamDiChaVit2024} $+$ &  & Full   & Partial &  Full   & Partial \\
    \midrule
    
     \simclr~\cite{chen2020simple,chen2020big}   & 70.72\footnotesize{$\pm$0.40} & 67.12\footnotesize{$\pm$0.39} & 56.96\footnotesize{$\pm$0.68} & \underline{64.44\footnotesize{$\pm$0.58}} & \underline{49.42\footnotesize{$\pm$0.63}} \\
     \simsiam~\cite{chen2021exploring}  & 70.44\footnotesize{$\pm$0.38} & 68.64\footnotesize{$\pm$0.56} & 57.72\footnotesize{$\pm$0.69} & 64.07\footnotesize{$\pm$0.31} & 48.52\footnotesize{$\pm$0.71} \\
     \ibot~\cite{zhou2021ibot}   & 70.71\footnotesize{$\pm$0.44} & 68.87\footnotesize{$\pm$0.51} & 57.81\footnotesize{$\pm$0.47} & 63.11\footnotesize{$\pm$0.32} & 47.84\footnotesize{$\pm$0.54} \\
     \dinotwo~\cite{oquab2024dinov}  & 70.03\footnotesize{$\pm$0.28} & 66.91\footnotesize{$\pm$0.52}  & 56.18\footnotesize{$\pm$0.66} & 63.42\footnotesize{$\pm$0.27} & 49.20\footnotesize{$\pm$0.72} \\
     MAE~\cite{he2022masked,kraus2024masked}  & \underline{70.27}\footnotesize{$\pm$0.78} & \underline{78.62}\footnotesize{$\pm$0.78} & \underline{64.21}\footnotesize{$\pm$0.73}  & 62.88\footnotesize{$\pm$0.49} & 47.76\footnotesize{$\pm$0.62} \\
    
    \midrule

     \textbf{\ourmaskabbr{}}  & \textbf{71.47}\footnotesize{$\pm$0.53}   & \textbf{84.02}\footnotesize{$\pm$0.49} &
\textbf{65.72}\footnotesize{$\pm$0.43} & \textbf{66.02}\footnotesize{$\pm$0.22} & \textbf{50.52}\footnotesize{$\pm$0.45}
  \\

    \bottomrule
  \end{tabular}
   \vspace{-2mm}

\end{table}

    
    




%% file: tables/segmentation.tex
\begin{table*}[t]
\centering
\setlength{\tabcolsep}{3.5pt}
\caption{\textbf{Comparing segmentation Performance on 38-Cloud~\cite{mohajerani2019cloud}.} We report the average and standard deviation from three runs. \ours{} outperforms all baselines in all metrics.}
\label{tab:segmentation}
\vspace{-2mm}
\begin{tabular}{lccccc}
    \toprule
    Model & Accuracy & IoU & Precision & Recall & F1 \\
    \midrule
    \channelvit~\cite{bao2024channel} & 0.945\scriptsize{$\pm$0.003} & 0.843\scriptsize{$\pm$0.012} & 0.919\scriptsize{$\pm$0.015} & 0.911\scriptsize{$\pm$0.004} & 0.915\scriptsize{$\pm$0.003} \\
    \dichavit~\cite{phamDiChaVit2024} & 0.951\scriptsize{$\pm$0.004} & 0.857\scriptsize{$\pm$0.011} & 0.924\scriptsize{$\pm$0.007} & 0.922\scriptsize{$\pm$0.006} & 0.923\scriptsize{$\pm$0.003} \\
    CA-MAE~\cite{kraus2024masked} & 0.946\scriptsize{$\pm$0.002} & 0.845\scriptsize{$\pm$0.003} & 0.917\scriptsize{$\pm$0.005} & 0.916\scriptsize{$\pm$0.003} & 0.916\scriptsize{$\pm$0.002} \\
    \dichavit~\cite{phamDiChaVit2024} $+$ CA-MAE~\cite{kraus2024masked} & 0.959\scriptsize{$\pm$0.001} & 0.886\scriptsize{$\pm$0.004} & 0.939\scriptsize{$\pm$0.005} & 0.943\scriptsize{$\pm$0.004} & 0.941\scriptsize{$\pm$0.002} \\
    \midrule
    \textbf{\oursbr} & \textbf{0.964\scriptsize{$\pm$0.001}} & \textbf{0.894\scriptsize{$\pm$0.002}} & \textbf{0.943\scriptsize{$\pm$0.002}} & \textbf{0.945\scriptsize{$\pm$0.001}} & \textbf{0.944\scriptsize{$\pm$0.001}} \\
    \bottomrule
\end{tabular}
\vspace{-4mm}
\end{table*}

%% file: tables/ours_ablation.tex
\begin{table}[h!]
  \caption{\textbf{\ours{} ablation study.} Replacing any component from \ours, such as substituting our \textit{DCP Masking} with the random patch masking from CA-MAE~\cite{kraus2024masked} ("w/o DCP Masking"), results in a performance drop. The decline is most significant when \textit{DCP Masking} is excluded. Incorporating all components consistently enhances performance across all three datasets.}
  \label{tab:main_ablation}
  \centering
  \begin{tabular}{l*{6}{c}}
    \toprule
    & \multicolumn{1}{c}{CHAMMI~\cite{chen2023chammi}} & \multicolumn{2}{c}{JUMP-CP~\cite{jumpcpSrinivas}} & \multicolumn{2}{c}{So2Sat~\cite{mediatum1483140}} \\
     \cmidrule(r){2-2} \cmidrule(r){3-4} \cmidrule(r){5-6} 
    Model     & Avg score  & Full   & Partial &  Full   & Partial \\
    \midrule
    \textbf{\ours} & \textbf{74.63} & \textbf{90.73}& \textbf{68.05} & \textbf{67.44} & \textbf{52.11}  \\
     w/o DCP Masking & 70.51 &  88.01 & 52.33  &  64.50 & 28.70
 
  \\
    w/o Hybrid Token Fusion      &  73.84 &  88.25 & 66.23    & 65.48 & 51.40 \\
    w/o Memory Tokens       &  73.62 &  87.81  & 67.21  &  65.18 & 50.46  \\
   
    w/o Channel-Aware Decoder      & 72.95 & 87.52& 67.05    & 65.78 & 49.88
  \\
    \bottomrule
  \end{tabular}
  \vspace{-2mm}
\end{table}

%% file: 6_analysis_and_discussion.tex
\input{tables/jumcp8_partial}

\input{tables/mask_strategies}

\noindent\textbf{Inference under varying channel configurations.} \cref{table:jumpcp8_partial} evaluates \ours{} and the best baseline \dichavit~\cite{phamDiChaVit2024} when tested on varying numbers of channels of JUMP-CP. We train the model using all eight channels and assess performance after removing channels, \eg, testing with seven channels, as shown in column "7," we averaged all $C^7_8 = 8$ possible combinations (refer to Appendix \cref{table:leave_one_channel_out}  for detailed results). \ours{} demonstrates enhanced robustness, particularly when some channels are missing during inference.

\noindent\textbf{Comparing masking strategies.}
\cref{tab:mask_strategies} compares various masking strategies applied to \ours. For fixed ratio approaches, we test several ratios (\eg, $50\%$, $75\%$) and report the best results.
\ourmaskabbr{} consistently outperforms others, achieving highest scores in both \textit{Full} and \textit{Partial} settings. Notably, while random patch masking is effective in \textit{Full}, it experiences a significant performance drop in \textit{Partial}. In contrast, dynamic channel masking methods adapted from \cite{bao2024channel,phamDiChaVit2024} provide better adaptability for partial settings, and our \ourmaskabbr{} approach improves even more. In addition, we analyze two variants of \methodfont{DCP} that we use in all our experiments: \methodfont{Combination} and \methodfont{Alternate}. The \methodfont{DCP Combination} unifies both patch and channel masks, while the \methodfont{DCP Alternate} switches between the two for each training iteration. Both variants outperform traditional patch- and channel-based masking, highlighting the effectiveness of joint channel-patch masking. Refer to \cref{sec:dcp_masking_more_analysis} for more analysis and hyperparameter settings for the \ourmaskabbr.

\begin{figure}[t]
  \centering
\includegraphics[width=1\linewidth]{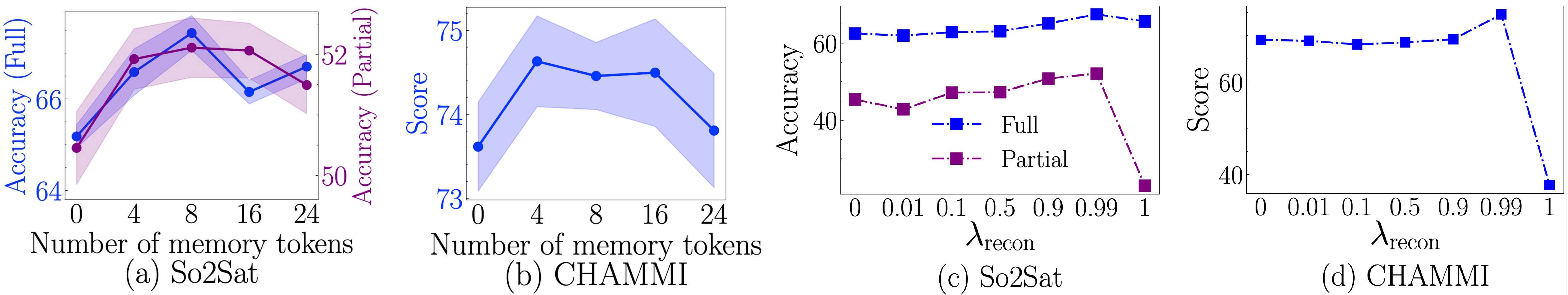}
\vspace{-6mm}
\caption{\textbf{Impact of the number memory tokens and reconstruction lambda $\lambda_{\mathrm{recon}}$ (\cref{eq:finalloss}).} \textbf{(a) \& (b)} Using $4-8$ tokens improves performance, however, using more memory tokens (\eg, $24$) may reduce the effectiveness. \textbf{(c) \& (d)} $\lambda_{\mathrm{recon}}=0$ means without the reconstruction loss, while $\lambda_{\mathrm{recon}}=1$ indicates only using the reconstruction loss. For $\lambda_{\mathrm{recon}}=1$ on So2Sat, we run linear probing. $\lambda_{\mathrm{recon}}=0.99$ works best for \ours{} on both datasets.}
\label{fig:num_memory_tokens_and_recon_lambda}
\end{figure}

\begin{figure}[t]
  \centering
\includegraphics[width=0.8\linewidth]{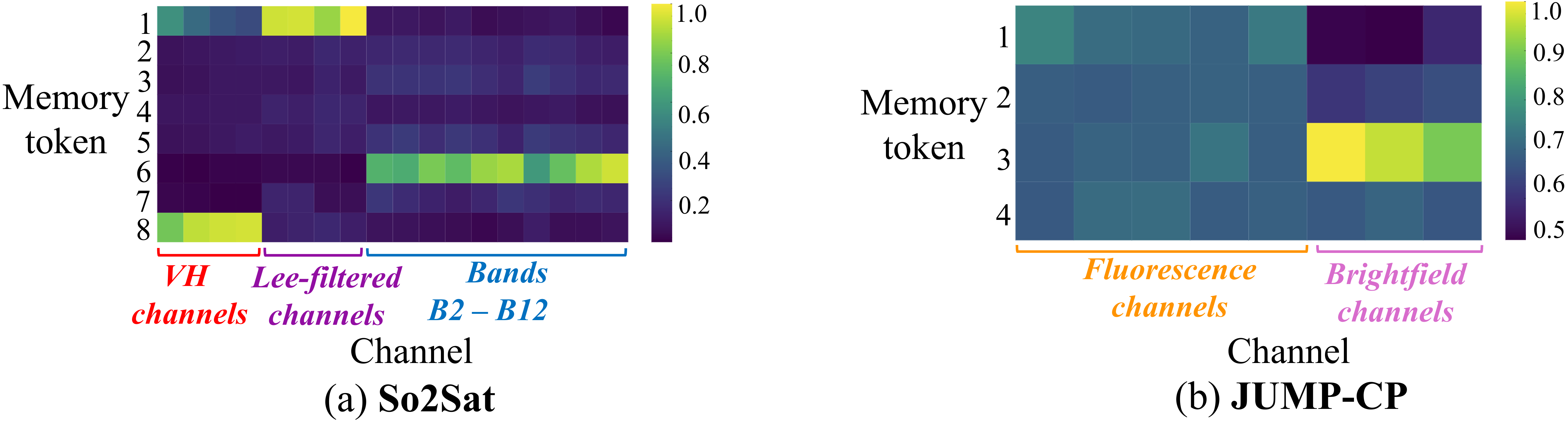}
\vspace{-2mm}
  \caption{\textbf{Attention between image patches and memory tokens of the encoder}. Each channel group focuses on different memory tokens. \textbf{(a) So2Sat:} \textit{VH} channels utilize memory token $8$, whereas \textit{Lee-filtered} channels attend more to memory token $1$. \textbf{(b) JUMP-CP:} \textit{Brightfield} channels focus on memory token $3$, while Fluorescence channels favor memory token $1$.}
\label{fig:attention_memory}
\end{figure}

\begin{figure}[h!]
  \centering
  \begin{minipage}[t]{0.5\textwidth}
    \centering
    \captionof{table}{\textbf{Comparison of token pooling methods.} \methodfont{Hybrid Token Fusion} achieves the best performance, demonstrating the benefit of leveraging both global and fine-grained local tokens.}
    \label{tab:hybrid_token_fusion}
    \vspace{2mm} 
    \setlength\tabcolsep{1pt} 
    \resizebox{\textwidth}{!}{
    \begin{tabular}{r*{3}{c}}
      \toprule
      & \multirow{2}{*}{CHAMMI} & \multicolumn{2}{c}{So2Sat} \\
      \cmidrule(r){3-4} 
      Token Pooling Method &  & Full & Partial \\
      \midrule
      $\mathrm{Avg}$ Patch tokens & 71.41 & 64.86 & 51.87 \\
      $\mathrm{[CLS]}$ token & 73.84 & 65.48 & 51.40 \\
      $\mathrm{[CLS]}$ + $\mathrm{Avg}$ Patch tokens & 73.96 & 66.23 & 49.45 \\
      \midrule
      \textbf{Hybrid Token Fusion (ours)} & \textbf{74.63} & \textbf{67.44} & \textbf{52.11} \\
      \bottomrule
    \end{tabular}
    }
  \end{minipage}\hfill
  \begin{minipage}[t]{0.46\textwidth}
    \centering
    \vspace{-1mm} 
    \includegraphics[width=\linewidth]{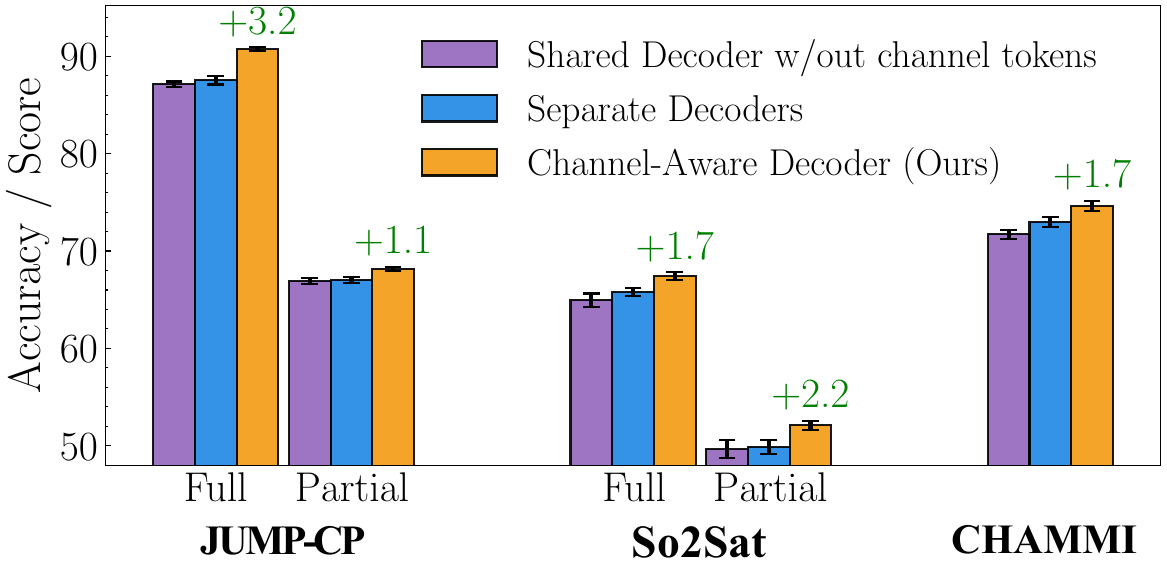}
    \caption{\textbf{Different Decoders when using with \ours}. Our Channel-Aware Decoder outperforms the best baseline by $1.1-3.2\%$ on all three datasets.}
    \label{fig:decoder_analysis}
  \end{minipage}
\end{figure}

\subsection{Model ablations and sensitivity analysis}
\noindent\textbf{Impact of Memory tokens.}
\cref{fig:num_memory_tokens_and_recon_lambda}(a) \& (b) show the accuracy achieved by numbers of \textit{memory tokens}.
Using memory tokens enhances performance, but beyond a certain point yields diminishing returns. This suggests that while memory tokens can improve performance, excessive reliance on them may negatively impact the interaction of patch features. We found that a default of $4$ memory tokens works well across the three datasets.

\noindent \textbf{Attention patterns of Memory tokens.}
\cref{fig:attention_memory} shows the attention patterns between image patches and memory tokens of the encoder. Each group of channels displays distinct preferences for specific memory tokens. \cref{fig:attention_memory}(a) shows that the \textit{VH} channels primarily focus on memory token $8$, while the \textit{Lee-filtered} channels have stronger attention toward memory token $1$. Similarly, for JUMP-CP in \cref{fig:attention_memory}(b), the \textit{Brightfield} channels allocate more attention to memory token $3$, whereas \textit{Fluorescence} channels show a slight preference for memory token $1$. This suggests that each type of channel utilizes different memory tokens to store the global information necessary for feature extraction.

\noindent\textbf{Token pooling strategies.}
\cref{tab:hybrid_token_fusion} compares the performance of various token pooling strategies on CHAMMI and So2Sat. In general, combining the $\mathrm{[CLS]}$ token with the average of the patch tokens gets better performance than either alone. By utilizing both the global $\mathrm{[CLS]}$ token and the fine-grained patch tokens, \methodfont{Hybrid Token Fusion} consistently outperforms others across all settings.

\noindent\textbf{Reconstruction loss lambda.}
\cref{fig:num_memory_tokens_and_recon_lambda}(c) \& (d) analyze the effect of the reconstruction weight $\lambda_{\mathrm{recon}}$ (\cref{eq:finalloss}) on model performance. Consistent with prior work~\cite{liang2022supmae,xin2024masked}, a large value of $\mathcal{L}_{\mathrm {recon}}$ gives the best results. However, performance drops significantly using just the MAE loss (\ie, $\lambda_{\mathrm{recon}}=1$), noting the contributions of both losses. In all our experiments, we set a fixed $\mathcal{L}_{\mathrm {recon}}$ to $0.99$.

\noindent\textbf{Channel-Aware Decoder analysis.}
\cref{fig:decoder_analysis} shows Channel-Aware Decoder in \ours{} outperforms both Separate Decoders~\cite{kraus2024masked} and Shared Decoder W/out Channel Tokens by  $1.1-3.2\%$ across three datasets in full and partial settings. Channel-Aware Decoder is also more efficient than Separate Decoders thanks to its shared parameters, \eg, $18X$ fewer parameters on So2Sat.

%% file: tables/jumcp8_partial.tex

 



\begin{table}[t]
\caption{\textbf{Performance on varying channel configurations during inference}. 
Columns report the \textit{mean}{\scriptsize{(std)}} across all channel combinations on JUMP-CP~\cite{jumpcpSrinivas}, 
\eg, "7" indicates testing on $7$ out of $8$ channels ($C_8^7=8$ combinations). 
\ours{} consistently shows improved robustness with missing channels at test time. 
Note that the reported std reflects variation across channel combinations, not model training variance. 
Refer to Table~\ref{table:leave_one_channel_out} in the Appendix for model variance report.}
\label{table:jumpcp8_partial}
\setlength{\tabcolsep}{2pt}
\centering
\begin{tabular}{lcccccccc}
\toprule 
&  \multicolumn{8}{c}{Number of channels at inference} \\
\cmidrule { 2 - 9 } 
Method & 8 & 7 & 6 & 5 & 4 & 3 & 2 & 1 \\
\midrule 

\dichavit{}~\cite{phamDiChaVit2024} 
& 69.19 
& 61.91\scriptsize{(9.3)} 
& 54.49\scriptsize{(12.4)} 
& 46.35\scriptsize{(13.4)} 
& 38.00\scriptsize{(12.4)} 
& 30.09\scriptsize{(9.3)} 
& 23.97\scriptsize{(4.9)} 
& 20.90\scriptsize{(1.6)} \\

\textbf{\ours} 
& \textbf{90.73} 
& \textbf{83.36}\scriptsize{(8.3)} 
& \textbf{74.55}\scriptsize{(11.7)} 
& \textbf{63.46}\scriptsize{(13.8)} 
& \textbf{50.85}\scriptsize{(13.9)} 
& \textbf{38.13}\scriptsize{(11.2)} 
& \textbf{27.62}\scriptsize{(6.4)} 
& \textbf{21.59}\scriptsize{(2.1)} \\
\bottomrule
\end{tabular}
\vspace{-4mm}
\end{table}

%% file: tables/mask_strategies.tex
\begin{table}[t]
  \caption{\textbf{Mask strategies in MCI when using with \ours}. While \textit{Random Patch} and \textit{Channel Sampling} perform similarly on Full channels, \textit{Random Patch} experiences a significant drop in Partial channel settings.  \ourmaskwithabbr{} demonstrates its effectiveness in both Full and Partial channel settings, outperforming all strategies across three datasets.}
  \label{tab:mask_strategies}
  \centering
  \setlength\tabcolsep{3pt}
    {
  \begin{tabular}{r*{6}{c}}
    \toprule
    & \multicolumn{1}{c}{CHAMMI} & \multicolumn{2}{c}{JUMP-CP} & \multicolumn{2}{c}{So2Sat} \\
     \cmidrule(r){2-2} \cmidrule(r){3-4} \cmidrule(r){5-6} 
    Mask Strategy     & Avg score  & Full   & Partial &  Full   & Partial \\
    \midrule

  Random Patch (fixed ratio) (\eg, ~\cite{kraus2024masked,bachmann2022multimae,liang2022supmae,kenyon2024vitally,eymael2024efficient}) & 70.51 & 88.01 & 52.33 & 64.50 & 28.70
 \\
 Random Patch (dynamic)  &  68.23  &  85.54 & 55.86  & 64.84 & 31.92  \\

  Channel (fixed ratio)~\cite{zhang2024multimodal} & 47.97 &  73.00 &
65.25 & 65.22 & 38.59
  \\
  Hierarchical Channel Sampling (dynamic, adapted \cite{bao2024channel}) & 69.86 & 83.71 & 67.75  & 65.20 & 48.77
 \\
  Diverse Channel Sampling (dynamic, adapted \cite{phamDiChaVit2024}) & 71.57 & 84.69 &
67.86  & 65.49 & 51.05 \\
  Channel $+$ Patch (fixed ratio)~\cite{wang2024hsimae,nedungadi2024mmearth} & 48.46 & 69.41 & 62.19 & 62.20 & 32.18 \\
  \midrule 
\textbf{\methodfont{DCP Combination}} ($\mathbf{p}_\mathbf{channel} = \mathbf{p}_\mathbf{patch} = \mathbf{0}$) (\textbf{ours}) & 73.75  &  85.95 & \textbf{68.36} & \textbf{67.44} & \textbf{52.11}
 \\
 \textbf{\methodfont{DCP Alternate}} ($\mathbf{p}_\mathbf{channel} = \mathbf{p}_\mathbf{patch} = \mathbf{0.5}$) (\textbf{ours}) &\textbf{74.63} & \textbf{90.73}
& 68.05 & 66.47 &
50.69\\
    \bottomrule
  \end{tabular}
  } 
\end{table}

%% file: 7_conclusion.tex
\section{Conclusion}
In this paper, we introduce \ours, a novel MAE-based method to enhance feature learning in Multi-Channel Imaging (MCI) ViTs. First, we introduce \ourmaskwithfont, in which we adaptively mask both image patches and channels and train the model to reconstruct the masked patches. Additionally, we incorporate \textit{memory tokens} to preserve global context across channels and \methodfont{Hybrid Token Fusion} module to combine features from local patch tokens and global class tokens. 
Furthermore, we propose \methodfont{Channel-Aware Decoder} to efficiently reconstruct channel-specific details. Experiments conducted on the CHAMMI, JUMP-CP, and So2Sat datasets demonstrate that \ours{} outperforms prior MCI-ViTs by \ourimprove, highlighting the significance of enhanced cross-channel interactions. 
Future research could explore the adaptation of the channel-aware framework to additional complex modalities, such as volumetric medical imaging, which is composed of sequential two-dimensional slices.

\newpage
\begin{ack}
This study was supported, in part, by the National Science Foundation under NSF-DBI awards 2134695 and 2134696. Any opinions, findings, and conclusions or recommendations expressed in this material are those of the author(s) and do not necessarily reflect the views of the supporting agencies.
\end{ack}

%% file: 8_appendix.tex

\clearpage
\appendix

\section{Broader Impacts and limitations} 
The development of \ours{} represents a significant advancement in Multichannel Imaging, with benefits such as improved medical diagnostics and faster healthcare research. Its applications in satellite imaging also hold potential for environmental monitoring. However, there are risks, including the possibility of misuse for invasive surveillance systems, highlighting the need for ethical considerations and responsible deployment.

While our model is capable of handling unseen channels by leveraging relationships learned from known ones (\cref{sec:novel_channels}), we have not thoroughly evaluated its performance in such scenarios. Adapting to novel channels requires mapping them to existing ones, which becomes more challenging under domain shifts. We leave a systematic investigation of this setting to future work. Additionally, our approach requires some extra hyper-parameter tuning, which can increase computational resource demands.

\section{Image patching}
\label{sec:patching_details}
Formally, let $ I \in \mathbb{R}^{h \times w \times c} $ represent a multi-channel input image, where $ (h, w) $ denotes the dimensions of the image and $ c $ is the number of channels. We begin by patchifying each channel into $n$ 2-D patches of size $p \times p$ in pixels. This process yields $n \times c$ patches, denoted as $\mathbf{x}_{i,j} \in \mathbb{R}^{p \times p}$, where $n = \frac{h \cdot w}{p^2}$ is the number of patches per channel, $i$ indicates the spatial location of a given patch, and $j$ denotes the $j$-th channel. Next, we flatten each patch $\mathbf{x}_{i,j} \in \mathbb{R}^{p^2}$, and then pass it into a shared linear projection $\mathbf{W} \in \mathbb{R}^{p^2 \times d}$. This results in patch tokens $\mathbf{t}_{i,j} = (\mathbf{x}_{i,j} \cdot \mathbf{W}) \in \mathbb{R}^{d}$, leading to a sequence of $n \times c$ patch tokens $[\mathbf{t}_{1,1}; \mathbf{t}_{2,1}; \ldots; \mathbf{t}_{n,c}]$. Note that we utilize a shared projection across all channels, thus the number of parameters remains constant regardless of the number of input channels.

\section{Combination of \dichavit{} with SSL} 
\label{sec:ssl_method_details}
We adopt the following SSL methods with \dichavit~\cite{phamDiChaVit2024} as baselines:  
\simclr~\cite{chen2020simple,chen2020big}, \simsiam~\cite{chen2021exploring}, 
\methodfont{MAE}~\cite{he2022masked}, \ibot~\cite{zhou2021ibot}, and \dinotwo~\cite{oquab2024dinov}. 

\simclr~\cite{chen2020simple,chen2020big} utilizes augmentation strategies that include random cropping, color distortion, and Gaussian blur to generate the dual views of an input image. This approach requires twice the computational resources per sample compared to single-path methods. 
Since hue and saturation are not well defined in multi-channel images, we apply the augmentation by only adjusting brightness and contrast. 

\simsiam~\cite{chen2021exploring} employs similar augmentations as \simclr, but it eliminates the use of negative pairs by stop-gradient. This operation helps prevent representation collapse by decoupling the optimization of the twin network branches, which simulates an alternating optimization process similar to the Expectation-Maximization (EM) approach. Additionally, \simsiam{} does not require momentum encoders.

\methodfont{MAE}~\cite{he2022masked} involves randomly masking a large portion of the input image and training the model to reconstruct the missing content. Unlike contrastive methods, \methodfont{MAE} does not depend on augmentation pipelines, which can be ineffective for heterogeneous channels in MCI \cite{roddan2024calibration,patnala2023generating}. Additionally, by processing only a small fraction of the image, this method significantly lowers computational demands, thereby enhancing efficiency. Note that combining \methodfont{MAE} with \dichavit~\cite{phamDiChaVit2024} results in a model very similar to \camae~\cite{kraus2024masked}, an extension of \methodfont{MAE} for multi-channel imaging. However, key components from \dichavit $-$ such as diverse regularizers, channel tokens, and the diverse channel sampling strategy $-$ are not present in \camae, making the combined model more expressive for handling multi-channel imaging.

\ibot~\cite{zhou2021ibot} combines masked image modeling with self-distillation by utilizing dual augmentation streams that include both masked and full views. This encourages the model to align representations across different views while reconstructing missing content. Particularly, \ibot{} uses clockwise masking introduced in \methodfont{BEiT}~\cite{bao2022beit}, where a block of image patches is masked each time. To adapt this clockwise masking for multi-channel images, we generate a mask for each channel, ensuring that the masked blocks only cover the areas in the channel. Compared to \methodfont{MAE}, \ibot{} introduces additional computational overhead due to its use of teacher-student distillation. Furthermore, the dual-stream processing increases the per-sample training cost, making \ibot{} more computationally demanding than single-view methods like \methodfont{MAEs}.
\smallbreak
\dinotwo~\cite{oquab2024dinov} employs multi-cropping alongside various augmentations, such as color jittering, Gaussian blur, and random solarization, as well as self-distillation techniques. Notably, \dinotwo{} also incorporates additional loss functions, such as \ibot{} loss~\cite{zhou2021ibot}  and \methodfont{KoLeo} loss~\cite{sablayrolles2018spreading}. During training, \dinotwo{} generates multiple crops from each image (\eg, $2$ global views and $8$ local views) which leads to significantly higher computational demands compared to other SSL methods. For example, training \dinotwo{} on JUMP-CP dataset using two GPUs requires approximately $6$ days, whereas \simclr{} takes around $2$ days and \methodfont{MAE} only requires $1$ day. Refer to \cref{fig:flops} for FLOPs counts.
\smallbreak
\noindent To integrate these SSL methods with \dichavit~\cite{phamDiChaVit2024}, we add the task loss and optimize it along with the SSL loss. Additionally, we explored another variant that included an additional branch utilizing standard augmentations recommended by the authors of the dataset to optimize the task loss and return the predictions at test time (\eg, only using random cropping, horizontal flipping, and thin-plate-spline transformation~\cite{donato2002approximate} for CHAMMI). We observed that relying exclusively on complex augmentation pipelines, like those used in \simclr, negatively impacts the performance of the combined model, while incorporating at least one view with regular augmentation enhances the model's learning capability. For self-distillation methods such as \dinotwo, we found that the performance of the teacher network is slightly better than that of the student network, thus we report the performance on the teacher network.

\section{\ourmaskwithabbr{} Algorithm}
\label{sec:mask_algorithm}
\cref{alg:mask_strategy} outlines the procedure of \methodfont{\ourmaskwithabbr{}} in \cref{sec:mask_strategy} of the main paper, where we combine both \textit{patch} and \textit{channel} masking strategies. In practice, setting \(\mathrm{p}_\mathrm{patch} = \mathrm{p}_\mathrm{channel} = 0\)  merges both patch and channel masks into a unified mask, and setting \(\mathrm{p}_\mathrm{patch} = \mathrm{p}_\mathrm{channel} = 0.5\)  allows the model to switch between patch and channel masks. We adopt these two straightforward configurations for all our experiments. For \(\mathrm{p}_\mathrm{patch} = \mathrm{p}_\mathrm{channel} = 0\) (\ie, \textit{DCP Combination}), we found that it works well with a small value of patch mask ratio $r_p$, and simply set $r_p = 0.25$ for all the experiments. For \(\mathrm{p}_\mathrm{patch} = \mathrm{p}_\mathrm{channel} = 0.5\) (\ie, \textit{DCP Alternate}), we used a larger value of $r_p = 0.75$. Refer to \cref{sec:dcp_masking_more_analysis} for more analysis on these hyperparameters.
\input{algorithm/algorithm_v2}

\section{Implementation details}
\label{sec:implementation_details_appendix}
 For most of the baseline models, such as \hypervit{}, \channelvit{}, \chadavit{}, and \dichavit{}, we utilize the implementation from \cite{phamDiChaVit2024}~\footnote{\url{https://github.com/chaudatascience/diverse_channel_vit}}. For \ibot{} and \dinotwo, we adapt the implementation from \cite{oquab2024dinov}~\footnote{\url{https://github.com/facebookresearch/dinov2}}. 
We employ a ViT small architecture with $21$M parameters as the backbone for all methods. To ensure that we maintain the same number of parameters as the baselines, we removed one layer from the encoder to incorporate it into the decoder for our method. For both CHAMMI and JUMP-CP, we use a patch size of $16$, while for the So2Sat dataset, we opt for a patch size of $8$. We trained the models using the AdamW optimizer, aiming to minimize the cross-entropy loss for the JUMP-CP and So2Sat datasets, while employing proxy loss for CHAMMI.
\smallskip

\noindent\textbf{CHAMMI dataset~\cite{chen2023chammi}.} For the baselines, we utilize the $\mathrm{[CLS]}$ token from the final layer as the feature representation and train the model to minimize the proxy loss~\cite{teh2020proxynca}. We then evaluate the model on various tasks using the evaluation code provided by ~\cite{chen2023chammi}, incorporating a 1-Nearest Neighbour classifier to calculate the macro-average F1-score for each task individually~\footnote{\url{https://github.com/broadinstitute/MorphEm}}. The channel-adaptive interfaces are adapted from the authors' implementation code~\footnote{\url{https://github.com/chaudatascience/channel_adaptive_models}}. In addition to the model, we apply the same data augmentation techniques recommended by the authors, including thin-plate-spline (TPS) transformations~\cite{donato2002approximate}. Each model is trained for $100$ epochs with a learning rate of $0.0004$ and a batch size of $64$.
\smallskip

\noindent\textbf{JUMP-CP~\cite{jumpcpSrinivas} and So2Sat~\cite{mediatum1483140} datasets}. Following \cite{bao2024channel,phamDiChaVit2024}, we warm up the learning rate for the first $10$ epochs, reaching a peak of $0.0004$. After this initial period, the learning rate gradually decays to $10^{-6}$ according to a cosine scheduler. To mitigate overfitting, we apply a weight decay of 0.04 to the weight parameters while excluding the bias and normalization terms.  We use the same data augmentation techniques as outlined in \cite{phamDiChaVit2024}. For the final prediction, we utilize $\mathrm{[CLS]}$ token from the Transformer encoder into a classifier head to predict the probability of each class. The final model checkpoint is selected based on validation performance. Each model is trained for $100$ epochs, with a batch size of $64$ on JUMP-CP and $128$ on So2Sat. 
\smallskip

\noindent\textbf{Compute resources.} Experiments conducted on So2Sat and CHAMMI utilize one NVIDIA RTX GPU with 48GB RAM, alongside three Intel(R) Xeon(R) Gold 6226R CPUs. For JUMP-CP experiments, two NVIDIA RTX GPUs and six Intel(R) Xeon(R) Gold 6226R CPUs were used.

\section{Additional Results and Analyses}
\label{sec:additional_results_appendix}

\subsection{FLOPs Count for Combination of DiChaViT with SSL}

\begin{figure}
  \centering
\includegraphics[width=0.5\linewidth]{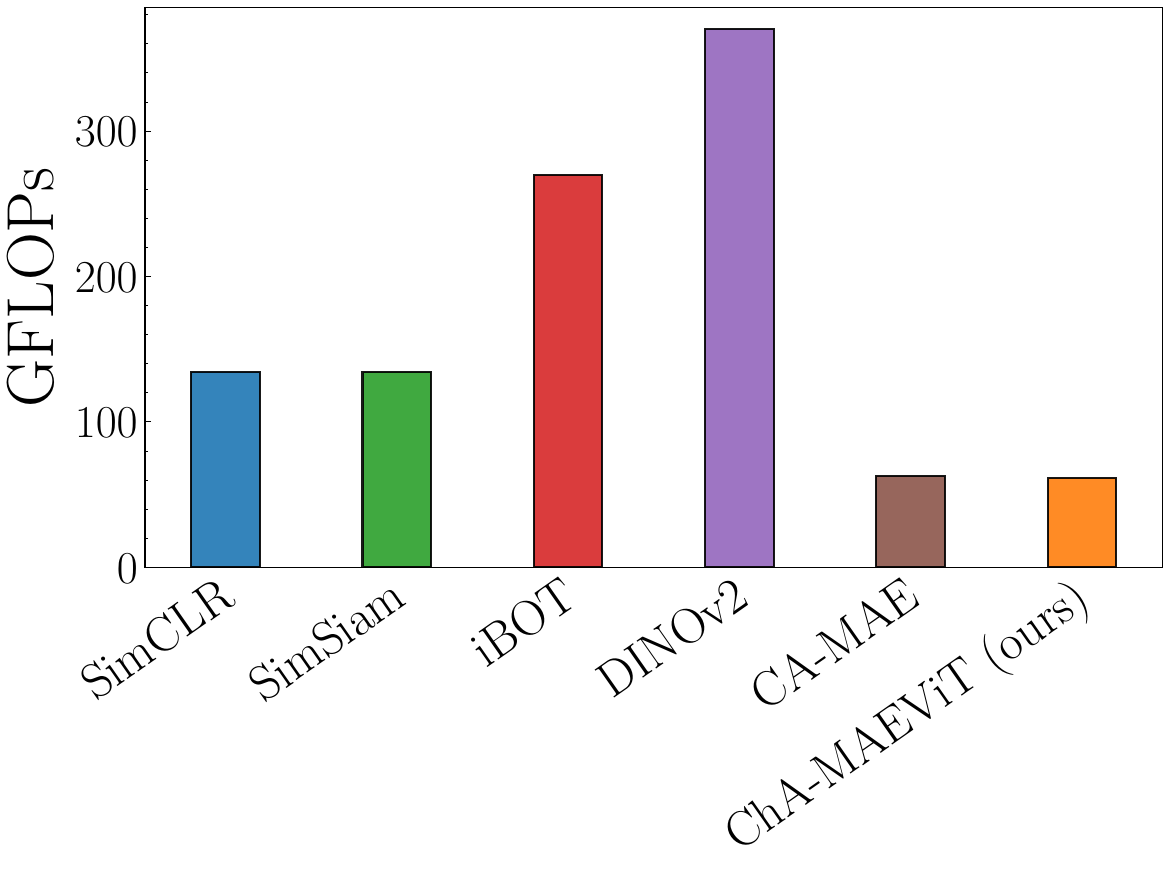}
  \caption{\textbf{FLOPs Count for Combination of DiChaViT with SSL}. We show FLOPs for each method when trained on JUMP-CP. \ours{} achieves the lowest FLOPs, $\approx1/6$ that of DINOv2.}
\label{fig:flops}
\end{figure}

In \cref{fig:flops}, we present the FLOPs for each of the SSL methods when combined with \dichavit{}, shown in \cref{tab:main_result_ssl} of the main paper. \ours{} achieves the lowest FLOPs, approximately one-sixth that of \dinotwo.

\subsection{CHAMMI Benchmark Results}
\input{tables/full_chammi_results}

\cref{tab:full_chammi_results} presents the F1 scores of various MCI-ViT models (ViT-S backbone) evaluated on the CHAMMI benchmark~\cite{chen2023chammi}. The evaluation consists of nine tasks, including six out-of-distribution (OOD) tasks. \ours{} demonstrates superior overall performance, achieving the highest scores in six out of the nine tasks across different settings.


\subsection{Performance when only using SSL Objectives}
\label{sec:ssl_result}
\begin{figure}
  \centering
\includegraphics[width=0.5\linewidth]{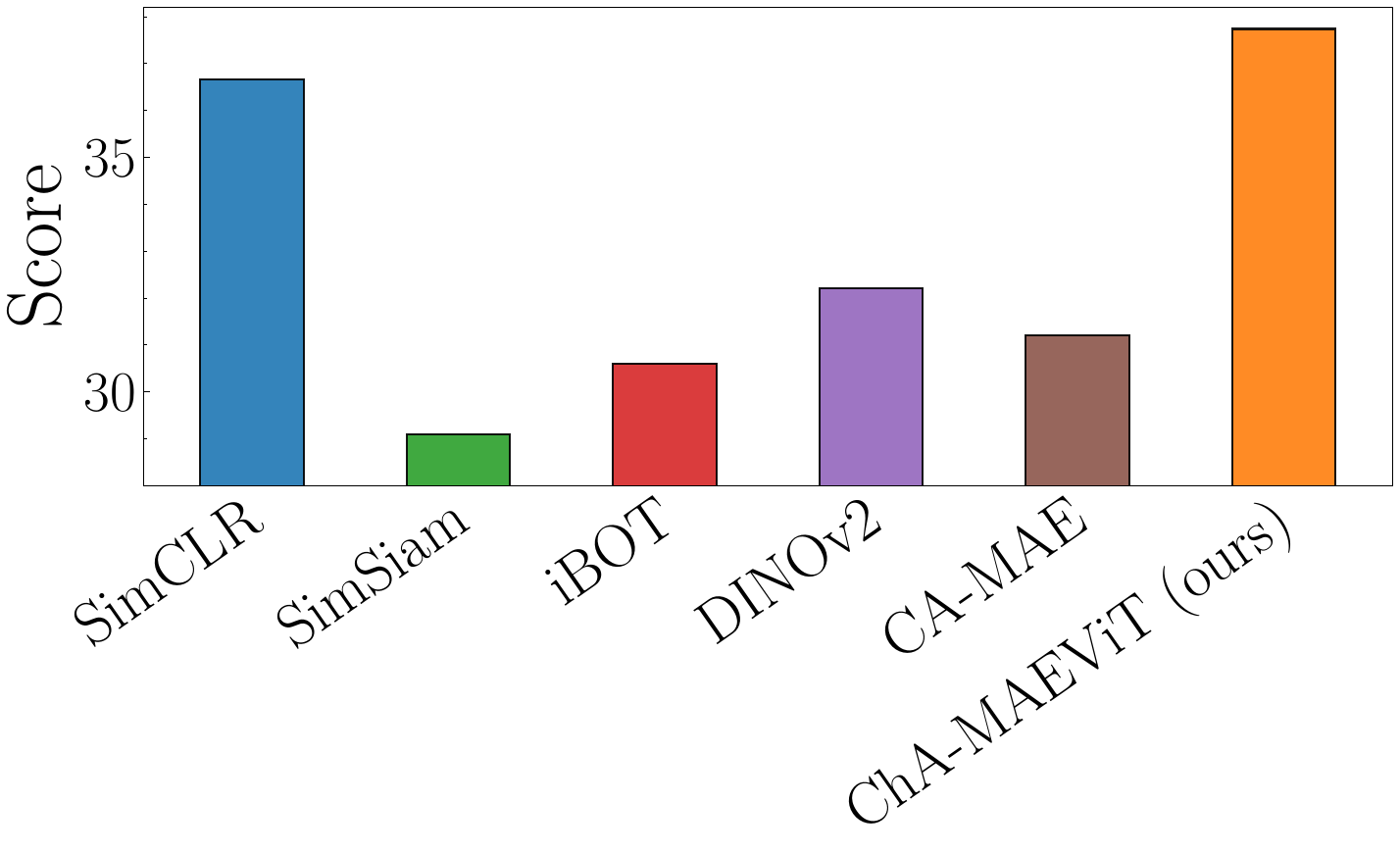}
  \caption{\textbf{Comparison of SSL methods on CHAMMI~\cite{chen2023chammi}}. All models were trained solely using SSL objectives, \ie, without any task loss. \ours{} achieves the highest score, outperforming other baselines by $1.0-8.5\%$, demonstrating the effectiveness of our approach over other methods in SSL settings for MCI.}
\label{fig:ssl_CHAMMI}
\end{figure}

\begin{figure}
  \centering
\includegraphics[width=0.8\linewidth]{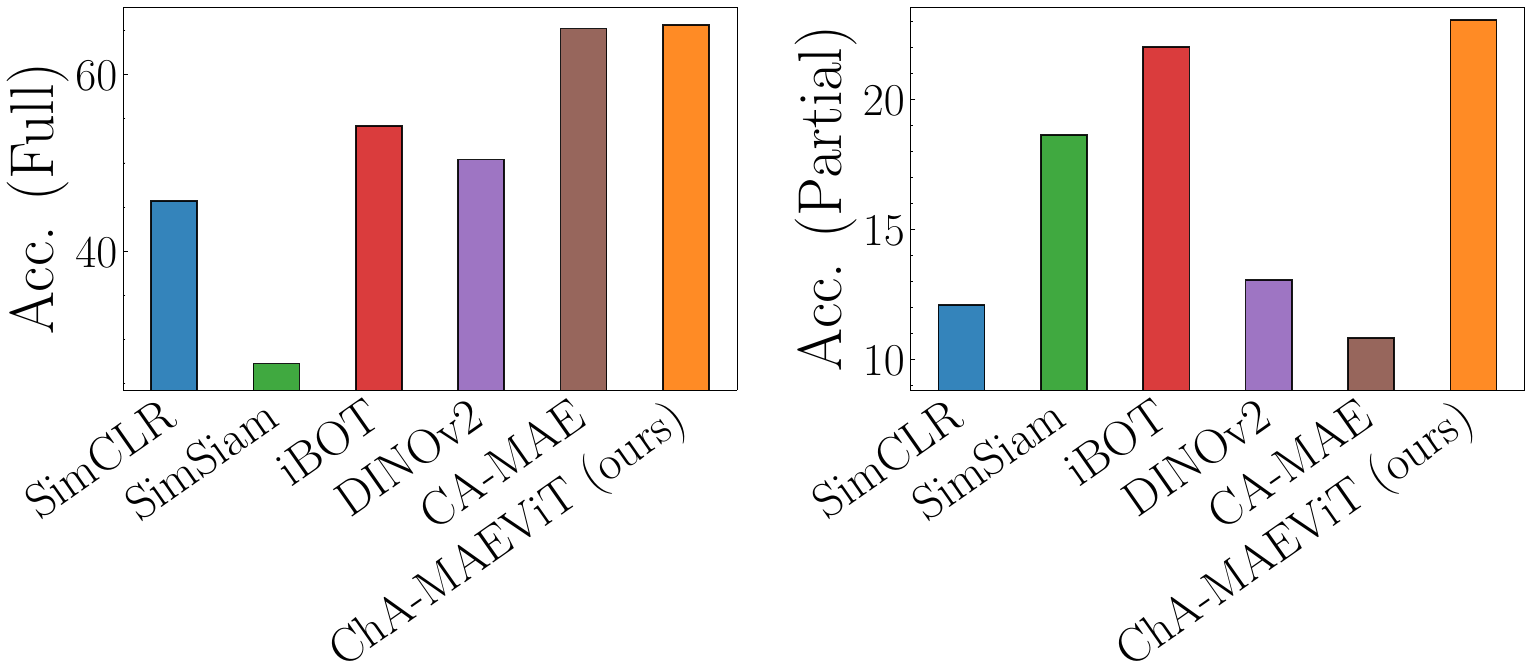}
  \caption{\textbf{Comparison of SSL methods on So2Sat~\cite{mediatum1483140}}. All models were trained solely using SSL objectives, \ie, without any task loss, then trained with linear probing for another $20$ epochs. \ours{} outperforms other baselines in both Full (left) and Partial channel settings (right), demonstrating its effectiveness for SSL in both scenarios.}
\label{fig:ssl_so2sat}
  \vspace{-2mm}
\end{figure}

\begin{figure}
  \centering
\includegraphics[width=0.8\linewidth]{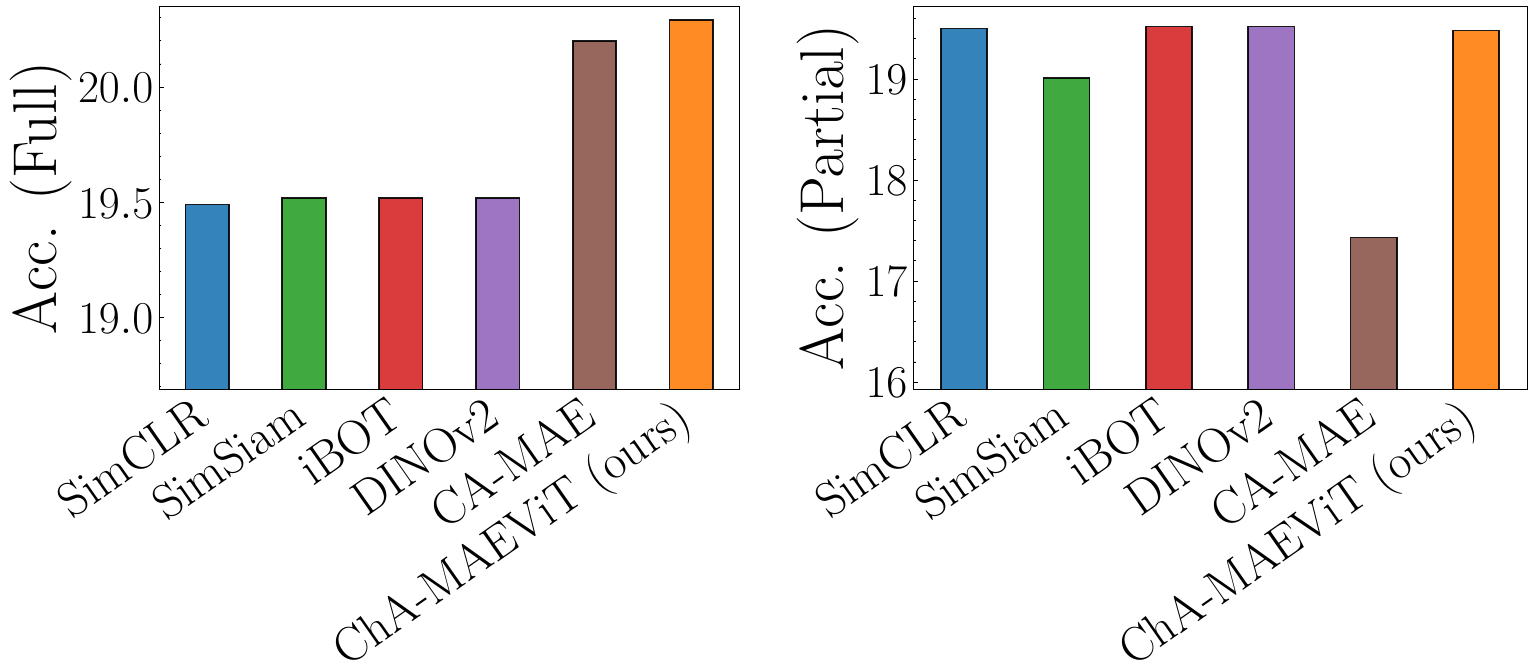}
  \caption{\textbf{Comparison of SSL methods on JUMP-CP~\cite{jumpcpSrinivas}}. All models were trained solely using SSL objectives, \ie, without any task loss, then trained with linear probing for another $20$ epochs. \ours{} achieves the highest score when tested on Full channels (left) and performs comparably to the top-performing baselines on Partial channel settings (right).}
\label{fig:ssl_jumpcp}
\end{figure}

We evaluate the effectiveness of \ours{} in the SSL setting by comparing the performance of different SSL methods. Specifically, all models were trained solely using SSL objectives, \ie, no task-specific loss was incorporated. For the So2Sat~\cite{mediatum1483140} and JUMP-CP~\cite{jumpcpSrinivas} datasets, we reported the accuracy from linear probing over $20$ epochs.
As shown in \cref{fig:ssl_CHAMMI}, \ours{} achieves the highest score of $37.7\%$, surpassing other approaches by $1.0-8.5\%$ on CHAMMI. In \cref{fig:ssl_so2sat}, our method consistently outperforms baseline models in both Full and Partial settings on So2Sat. A similar trend is observed for JUMP-CP in \cref{fig:ssl_jumpcp}.
This demonstrates the effectiveness of our method in SSL settings for MCI, highlighting its ability to extract meaningful representations in scenarios where supervised labels are unavailable.

\subsection{Leave-One-Channel-Out at Test Time}
\input{tables/leave_one_channel_out}

In \cref{table:leave_one_channel_out}, we present the results of training the model using all eight channels of JUMP-CP and then testing it with various combinations of seven channels. This provides a detailed result for column "7" of \cref{table:jumpcp8_partial} in the main paper, which represents \(C^7_8=8\) different channel combinations. For each combination, we report the mean and standard deviation of the model's performance based on three runs. Our results demonstrate that \ours{} achieves an improvement of $17-23\%$ for each combination compared to the baseline models \dichavit~\cite{phamDiChaVit2024} and \channelvit~\cite{bao2024channel}.

\subsection{\ourmaskwithabbr{} Analysis}
\label{sec:dcp_masking_more_analysis}
\begin{figure}
  \centering
\includegraphics[width=0.6\linewidth]{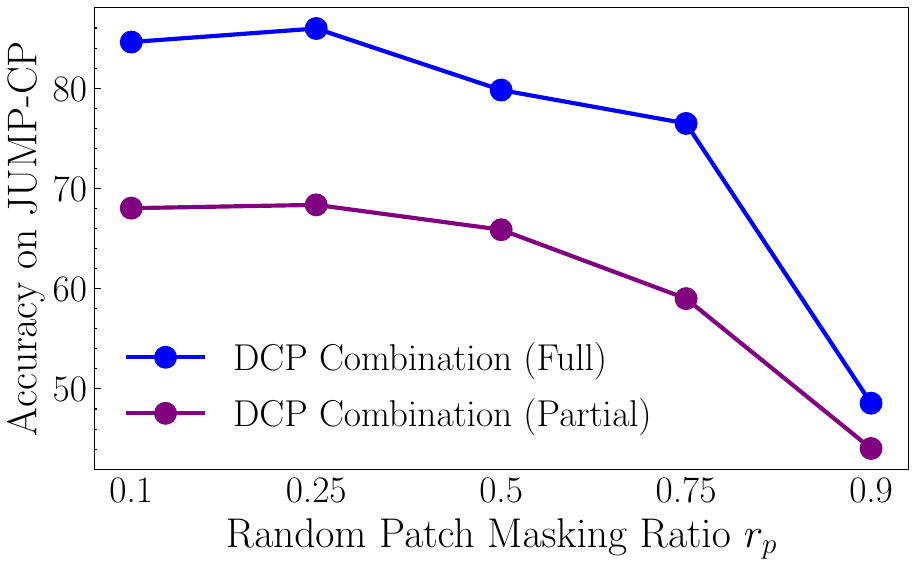}
 \caption{\textbf{\textit{DCP Combination} with varying masking ratios $r_p$}. We train \ours{} using a combined mask integrating both channel- and patch-level masking, \ie, \textit{DCP Combination} ($\mathrm{p}_\mathrm{channel} = \mathrm{p}_\mathrm{patch} = 0$). The blue and purple lines depict accuracy on JUMP-CP for Full and Partial channel settings, respectively, across different patch masking ratios $r_p$.
 }
\label{fig:our_mask_jumpcp}
\end{figure}

\noindent \textbf{\textit{DCP Combination} with varying masking ratios $r_p$}. \cref{fig:our_mask_jumpcp} shows the impact of varying \textit{random patch masking ratios} ($r_p$) on the performance of \ours{} when trained with \methodfont{DCP Combination} ($\mathrm{p}_\mathrm{patch} = \mathrm{p}_\mathrm{channel} = 0$). The blue and purple lines indicate the performance for full and partial channels on JUMP-CP, respectively. As the masking ratio $r_p$ increases, accuracy for both full and partial channels decreases due to excessive information loss, making reconstruction more difficult. Since \methodfont{DCP Combination} masks both patches and channels, it is reasonable to use a small $r_p$, in contrast to previous studies that often utilize a high ratio (\eg, $0.75$). For example, a $r_p$ of $0.25$, when combined with dynamic channel masking, together mask $\approx 0.65$ patches in the original images. We found \textit{DCP Combination} works well with a small value of $r_p$, thus we use $r_p$ of $0.25$ as default value for all the datasets for \textit{DCP Combination}.

\begin{figure}
  \centering
\includegraphics[width=0.65\linewidth]{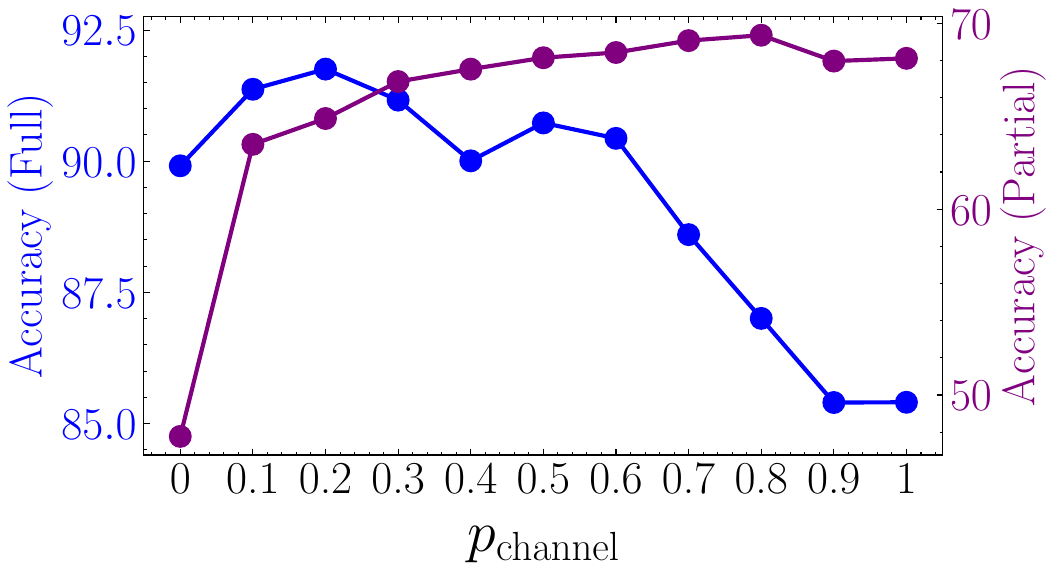}
 \caption{\textbf{\textit{DCP Alternate} with different proportions of $\mathbf{p}_\mathbf{channel}$ and $\mathbf{p}_\mathbf{patch}$}. To evaluate the effects of \textit{channel-} and \textit{patch-} level masks in \ours{} during alternating masking, we vary $\mathrm{p}_\mathrm{channel}$ while setting $\mathrm{p}_\mathrm{patch} = 1 -\mathrm{p}_\mathrm{channel}$.  $\mathrm{p}_\mathrm{channel}=1$ indicates only using channel-level masking, whereas $\mathrm{p}_\mathrm{channel}=0$ indicates only using patch-level masking. The blue and purple lines represent accuracy on JUMP-CP for Full and Partial channel settings respectively, across different values of $\mathrm{p}_\mathrm{channel}$. 
 }
\label{fig:p_channel_vs_p_patch_JUMPCP}
\end{figure}

\smallbreak
\noindent \textbf{\textit{DCP Alternate} with varying proportions of \(\mathbf{p}_\mathbf{channel}\) and \(\mathbf{p}_\mathbf{patch}\)}. \cref{fig:p_channel_vs_p_patch_JUMPCP} evaluates the performance of \textit{DCP Alternate} in \ours{} by varying the proportions of \textit{channel}- and \textit{patch}-level masking. We adjust \(\mathrm{p}_\mathrm{channel}\), while setting \(\mathrm{p}_\mathrm{patch} = 1 - \mathrm{p}_\mathrm{channel}\) to control the contribution of each masking strategy. For example, \(\mathrm{p}_\mathrm{channel} = 1\) indicates exclusive channel-level masking, and \(\mathrm{p}_\mathrm{channel} = 0\) denotes only using patch-level masking. The blue and purple lines represent accuracy on JUMP-CP for Full and Partial channel settings, respectively, across different \(\mathrm{p}_\mathrm{channel}\) values. 
We observed that using both channel and patch masks together improves performance. Additionally, a higher channel-level masking proportion (\ie, larger \(\mathrm{p}_\mathrm{channel}\)) improves accuracy in the Partial setting (purple) but leads to a decline in the Full setting (blue), highlighting the trade-off between these masking strategies. For \textit{DCP Alternate}, we found that $r_p=0.75$ works well across experiments.

\begin{figure}
  \centering
\includegraphics[width=0.7\linewidth]{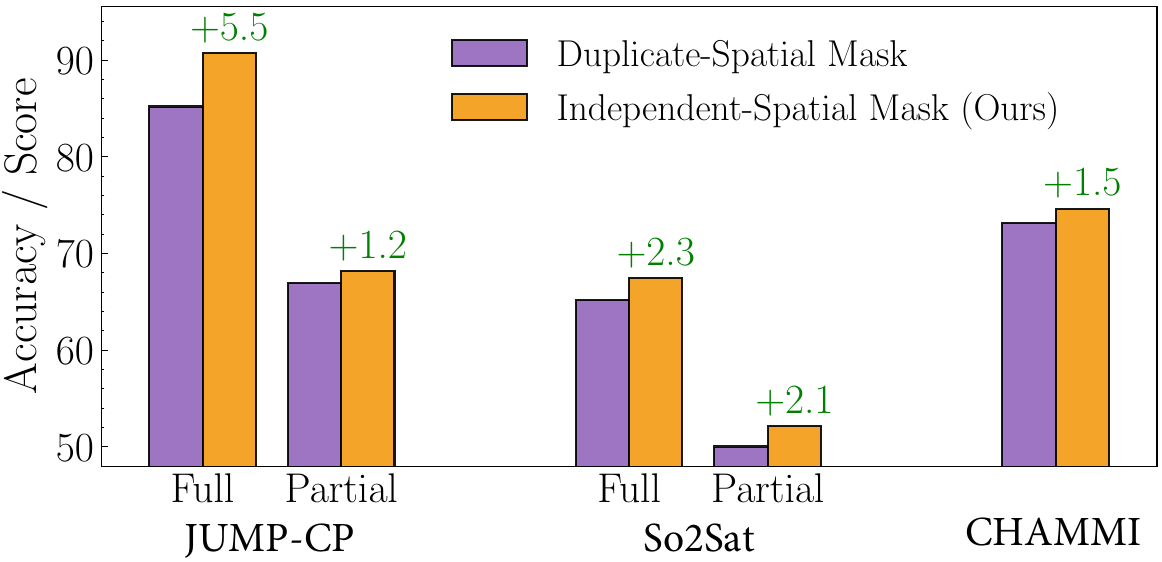}
 \caption{\textbf{Patch Masking Strategies in DCP}. We observed that using a different mask for each channel outperforms duplicating the same mask across all channels (purple), resulting in improvements of $1.2-5.5\%$ across three datasets. 
 }
\label{fig:patch_mask_ablation}
\end{figure}

\smallbreak
\noindent \textbf{Patch Masking Strategies in DCP}. \cref{fig:patch_mask_ablation} compares two patch masking strategies with DCP. The first strategy, \textit{Duplicate-Spatial Mask}, creates a patch mask for one channel and then duplicates the mask across all other channels (\eg, ~\cite{nedungadi2024mmearth,wang2024hsimae}). The second approach, \textit{Independent-Spatial Mask}, generates a patch mask for each channel independently, resulting in varied patch positions. We set the same masking ratio for both strategies. Our findings indicate that using a different mask for each channel performs better than the duplicating one, yielding an improvement of $1.2-5.5\%$ across three datasets.

\subsection{Impact of memory tokens on reconstruction loss} 
\label{sec:memory_token_loss_appendix}
\begin{figure}
  \centering
\includegraphics[width=0.7\linewidth]{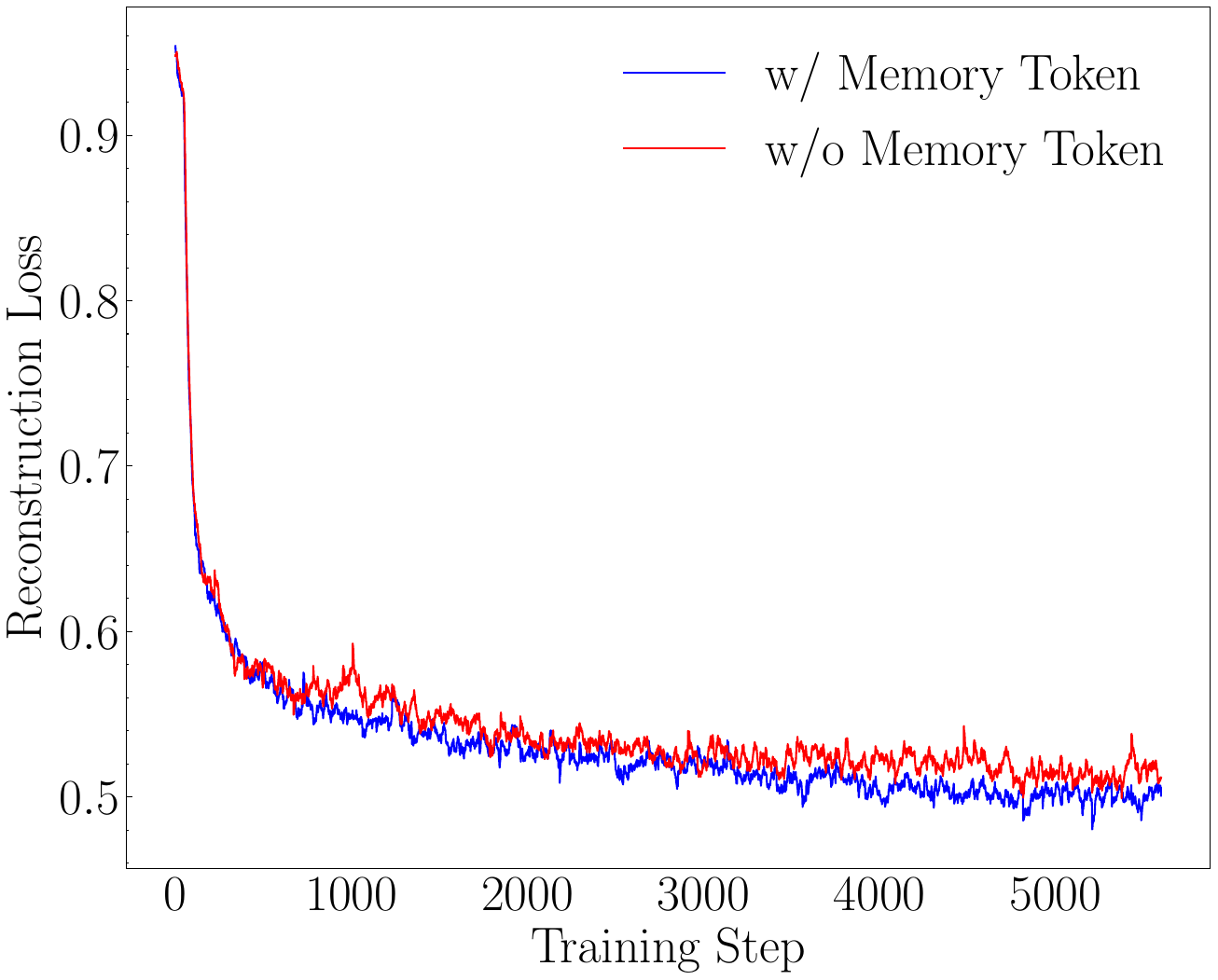}
  \caption{\textbf{Impact of memory tokens on the reconstruction loss}. Using memory tokens helps the reconstruction task, as demonstrated by the lower reconstruction loss observed on So2Sat.}
\label{fig:reconstruction_loss_so2sat}
\end{figure}

\cref{fig:reconstruction_loss_so2sat} shows the effect of memory tokens on reconstruction loss during training. Evaluated on the So2Sat dataset, the model using memory tokens (blue line) achieves lower and smoother loss compared to those without them (red). This indicates that memory tokens may improve the model's ability to capture global information, helping to ease the learning process.

\subsection{Number of Decoder Blocks of \textbf{\ours}}
\begin{figure}
  \centering
\includegraphics[width=0.6\linewidth]{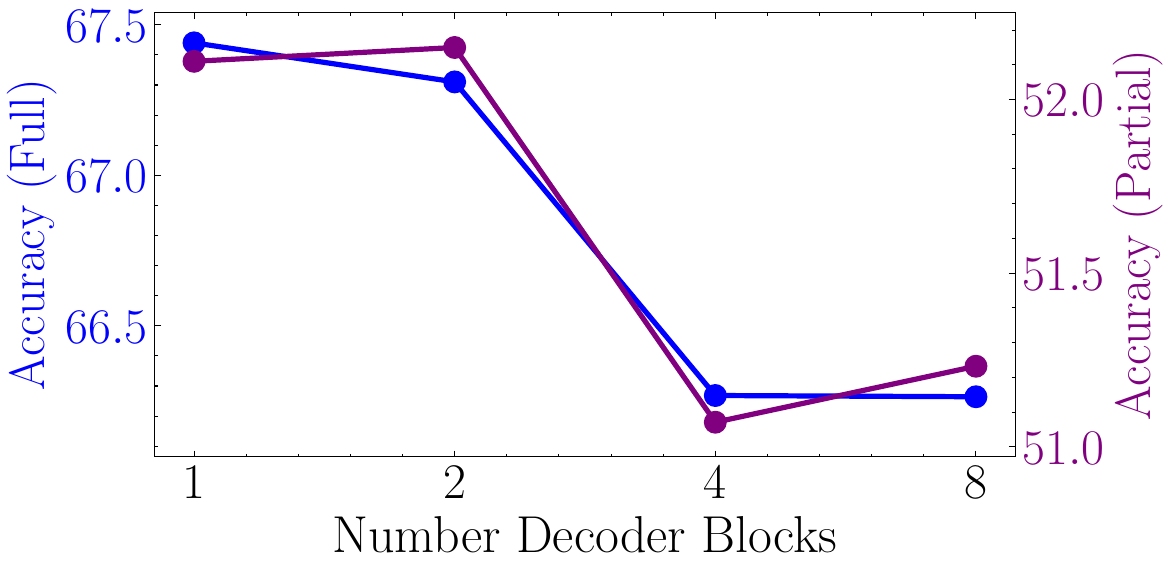}
  \caption{\textbf{Number of Decoder Blocks of \ours{}}. We show the performance of \ours{} when trained on So2Sat with different numbers of blocks in the decoder. We observed that using $1-2$ blocks gives the best performance.}
\label{fig:num_decoder_blocks}
\end{figure}

\cref{fig:num_decoder_blocks} illustrates the performance of \ours{} when trained on the So2Sat dataset with varying numbers of decoder blocks. We found that a configuration of 1 to 2 blocks yields the best performance.

\subsection{Scaling Model Size}
\label{sec:different_model_size_analysis}
\input{tables/bigger_model}

\cref{tab:scale_bigger} compares the performance of \ours{} with the two strongest supervised baselines: \channelvit~\cite{bao2024channel} and \dichavit~\cite{phamDiChaVit2024}. To evaluate the impact of model scaling, each model is evaluated using two backbone architectures: ViT-S ($21$M parameters) and ViT-B ($85$M parameters). Larger models exhibit improved performance, particularly on CHAMMI and JUMP-CP. Our approach consistently outperforms the baselines across all three datasets, regardless of the backbone used. In \ours, we reallocate one Transformer block from the encoder to the decoder, ensuring that all models maintain the same parameter count.

\subsection{Reconstruction Images}
\label{sec:qualitative_reconstruction_analysis}

\begin{figure*}
  \centering
\includegraphics[width=1\linewidth]{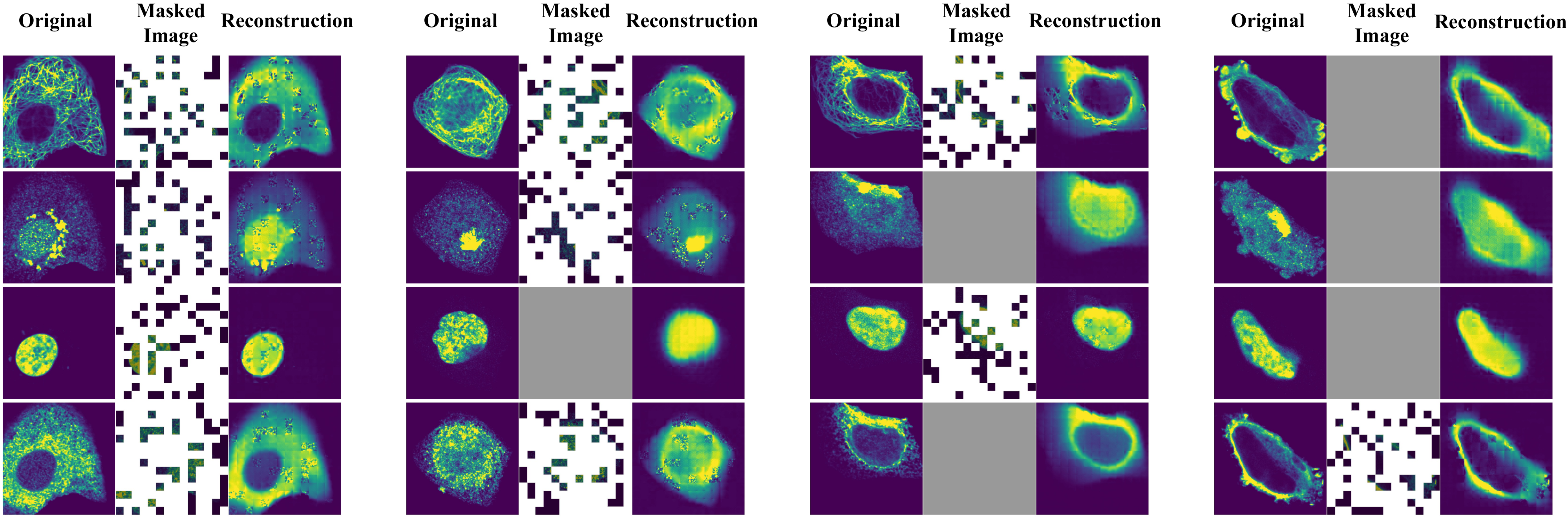}
  \caption{\textbf{Visualization of original, masked, and reconstructed images on CHAMMI-HPA~\cite{chen2023chammi}, which contains 4 channels}. We show the reconstruction of \ours, utilizing a ViT-S/16 backbone. \ours{} demonstrates its capability to reconstruct masked channels and patches under different channel masking settings ($0-3$ masked channels with a fixed \textit{patch masking ratio} $r_p=75\%$). Gray blocks indicate fully masked channels.}
\label{fig:recon_img}
  \end{figure*}
  
\cref{fig:recon_img} shows the \textit{original}, \textit{masked}, and \textit{reconstructed} images from the CHAMMI-HPA dataset~\cite{chen2023chammi}, which contains four channels. Each set of images shows the original image on the left, followed by the masked image created using \ourmask{} (with patch masking ratio \(r_p=75\%\)), and finally, the reconstructed image produced by \ours{} (utilizing a ViT-S/$16$ backbone) on the right. The gray blocks indicate the entire channel has been masked. \ours{} demonstrates its ability to leverage cross-channel dependencies as it can reconstruct the entire channels using available patches from the other channels. 

Note that the reconstructions lack some high-frequency details, as the primary goal of \ours{} is to learn robust, high-quality representations rather than producing detailed reconstructions. Our approach prioritizes the encoder's ability to capture high-level semantic and cross-modal relationships, with a deliberately lightweight decoder designed to validate the encoding of essential latent information. Additionally, following prior MAEs (\eg, ~\cite{he2022masked,kraus2024masked,noman2024rethinking,reed2022scale}), we employ the mean squared error (MSE) loss for reconstruction, which often results in blurry outputs by averaging plausible solutions~\cite{reed2022scale,mustafa2022training}. This leads to reconstructions dominated by low-frequency components, an expected outcome of the design choice.

\subsection{Novel Channels at Inference}
\label{sec:novel_channels}
For the novel channels setting, we trained \ours{} using the first three channels (\ie, channels $0, 1, 2$) and then tested it on two other channels (\ie, channels $3, 4$) of JUMP-CP. One of the main challenges is that new channels do not have learned \textit{channel tokens}. To address this, we assume that some unlabeled test images are available at test time. We initialize the channel tokens for the new channels randomly and then fine-tuned these channel tokens on the unlabeled test images using the reconstruction loss in \cref{eq:reconloss}. Throughout this process, we kept the entire model frozen and only fine-tuned the newly initialized channel tokens.

\begin{figure}
  \centering
\includegraphics[width=0.5\linewidth]{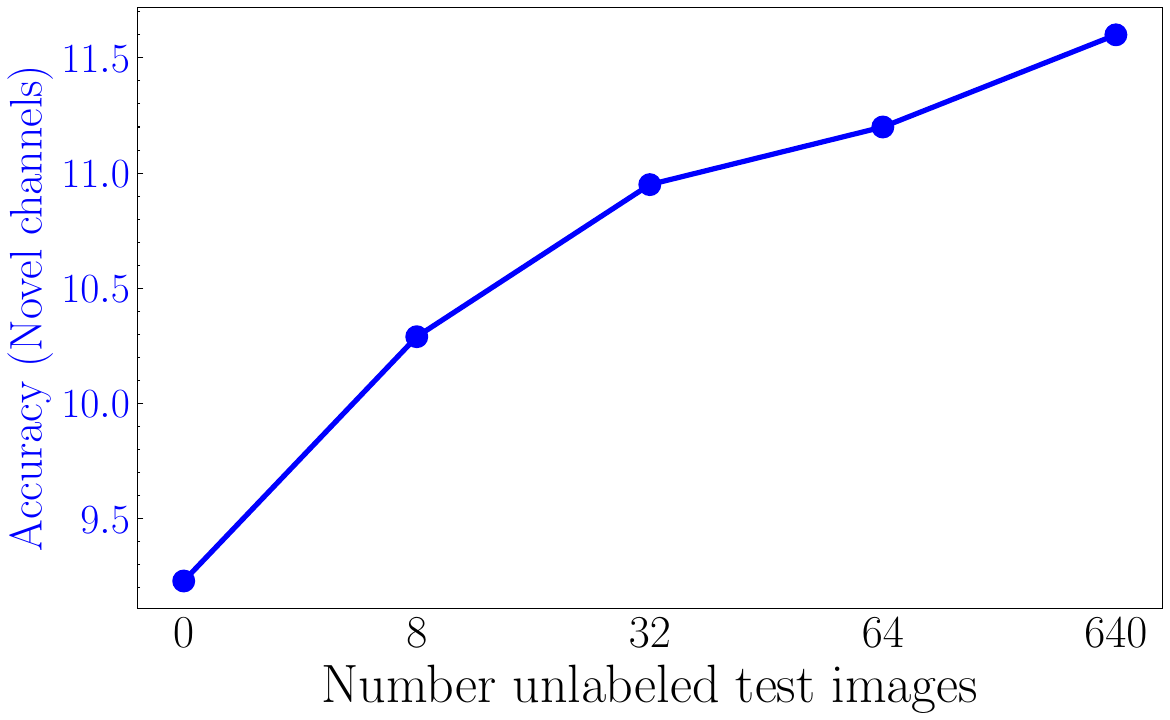}
  \caption{\textbf{Novel Channels at Inference}. Fine-tuning the \textit{channel tokens} of novel channels with unlabeled test images boosted performance from $9.2\%$ (without fine-tuning, denoted as "0") to $11.6\%$ (with fine-tuning on 640 unlabeled test images).}
\label{fig:new_channels}
\end{figure}

In \cref{fig:new_channels}, we illustrate the accuracy on novel channels with varying numbers of unlabeled test images. For example, "0" indicates that no fine-tuning was performed, while "8" represents that eight test images (each containing only the two novel channels) were used to fine-tune the channel tokens before testing the model on the remaining test set. We observed that fine-tuning the channel tokens with some unlabeled test images improved performance from $9.2\%$ (without fine-tuning) to $11.6\%$ (with fine-tuning on 640 unlabeled test images). Additionally, to make an upper performance bound, we trained \ours{} on the training set of the novel channels (\ie, channels $3$ and $4$). This model achieved an accuracy of $46.1\%$, indicating there is still a significant performance gap, which we leave for future work.

%% file: algorithm/algorithm_v2.tex
\begin{algorithm}[t]
\small
\caption{\ourmask}
\label{alg:mask_strategy}
\Input{Number of patches per channel $n$;\\
Number of channels $c$;\\
Patch mask ratio $r_p$;\\
Probability of using patch masking $\mathrm{p}_\mathrm{patch}$;\\
Probability of using channel masking $\mathrm{p}_\mathrm{channel}$;\\
\quad $ (0 \leq \mathrm{p}_\mathrm{patch} + \mathrm{p}_\mathrm{channel} \leq 1$)\\
}

\tcp{Random Patch Mask} 
$\mathbf{p\_mask} = \mathbf{0}^{n \times c}$ \quad \textit{// initialize patch mask}\\
{\bf For $j=1, 2,\dots, c$}: \\
\quad $\mathbf{p\_mask}[:, j] = \mathrm{generate\_random\_patch_\_mask}(n, r_p)$ \\
{\bf EndFor}\\

\tcp{Dynamic Channel Mask} 
Uniformly sample a number of channels to mask  $k \sim \mathcal{U}\{0, 1, \ldots, c-1\}$ \\
Uniformly sample $k$ masked channels 
$\mathcal{C}' = \mathrm{Sample}(\{1, \dots, c\}, k)$\\
$\mathbf{c\_mask} = \mathbf{0}^{n \times c}$ \quad \textit{// initialize channel mask}\\
{\bf For $j \in \mathcal{C}'$}:\\
    \quad $\mathbf{c\_mask}[:, j] = \mathbf{1}^n$ \quad \textit{// mask whole channel $j$}\\ 
{\bf EndFor}\\
\tcp{Masks For Current Iteration}
Sample $s \sim \mathcal{U}(0, 1) \in \mathbb{R}$ \\
\textbf{If} $s < \mathrm{p}_\mathrm{patch}$:\\
\quad $\mathbf{mask} = \mathbf{p\_mask}$ \tcp{Using patch mask}
\textbf{Else If} $s < \mathrm{p}_\mathrm{patch} + \mathrm{p}_\mathrm{channel}$:\\
\quad $\mathbf{mask} = \mathbf{c\_mask}$ \tcp{Using channel mask}
\textbf{Else}: \tcp{Combining both masks}
\quad $\mathbf{mask} = \mathbf{p\_mask} \lor \mathbf{c\_mask}$ \\

\Output{$\mathbf{mask} \in \{0, 1\}^{n \times c}$, where $1$ denotes mask}
\end{algorithm}

%% file: tables/full_chammi_results.tex
\begin{table*}[t]
 \caption{\textbf{F1 Scores of MCI-ViT models on CHAMMI benchmark}~\cite{chen2023chammi}. \ours{} demonstrates better overall performance on CHAMMI, achieving the highest scores in $6$ out of $9$ tasks across various settings. "OOD" refers to out-of-distribution tasks.  All models have the same number of parameters, except \camae{} with $3X$ more parameters due to its use of separate decoders.}
  \label{tab:full_chammi_results} 
  \centering
  \setlength\tabcolsep{2pt}
  \resizebox{\textwidth}{!}{ 
  \begin{tabular}{r*{13}{c}}
    \toprule
    & \multicolumn{4}{c}{Average OOD} 
    & \multicolumn{2}{c}{WTC} 
    & \multicolumn{3}{c}{HPA} 
    & \multicolumn{4}{c}{CP}   \\
    \cmidrule(r){2-5} \cmidrule(r){6-7} \cmidrule(r){8-10} \cmidrule(r){11-14}
    Model & Mean & WTC & HPA & CP & Task1 & Task2 & Task1 & Task2 & Task3 & Task1 & Task2 & Task3 & Task4 \\
    \midrule

    \hypervit~\cite{haHypernetworks2016} & 47.17	&45.78	&67.61&	28.11	&58.83&	45.78&	88.78&	82.70&	52.52	&82.13&	53.74	&23.16&	7.42 \\
    
 \depthwisevit~\cite{chen2023chammi} & 50.44 & 52.19 & 71.41 & 27.72 & 69.81 & 52.19 & 91.65 & \textbf{88.04} & 54.78 & 81.24 & 54.08 & \textbf{23.21} & 5.87 \\

 \methodfont{TempMixingViT}~\cite{plummerNPAS2022,savarese2018learning,phamMixtureGrowth2024} & 47.33 & 51.52 & 64.80 & 25.66 & 61.66 & 51.52 & 85.01 & 79.91 & 49.69 & 77.45 & 48.83 & 22.56 & 5.60 \\
     
\camae{}~\cite{kraus2024masked} $+$ Sup. Loss & 48.82 & 61.85 & 56.45 &\textbf{28.15} & 77.57 & 61.85 & 84.45 & 71.89 & 41.01 & 76.49 & 56.52 & 19.43 & \textbf{8.49} \\

  \chadavit~\cite{chadavit} &   50.82 & 67.18 & 60.67 & 24.60 & 77.58 & 67.18 & 87.49 & 75.94 & 45.41 & 83.92 & 45.58 & 21.94 & 6.28 \\

    \channelvit~\cite{bao2024channel} & 52.54 & 67.58 & 62.35 & 27.81  & 78.36 &  67.58 &  83.93 &  76.73 &  47.97 &  77.70 &  55.16 &  21.89 &  6.38  \\

    \dichavit~\cite{phamDiChaVit2024}  & 55.36 &  75.18 & 64.36 & 26.53 	&80.87 & 75.18 & 88.08 & 79.26 & 49.45 & 84.08 & 53.03 & 20.95 & 5.60\\
    \midrule
    \textbf{\oursbr}     & \textbf{58.02} & \textbf{77.15} & \textbf{72.11} & 24.81 & \textbf{84.52} & \textbf{77.15} & \textbf{94.14} & 87.47 & \textbf{56.75} &  \textbf{90.89} & \textbf{56.68} & 10.25 & 7.50 \\
    \bottomrule

  \end{tabular}
  }
  
\end{table*}

%% file: tables/leave_one_channel_out.tex
\begin{table*}[tbp]
\caption{\textbf{Analysis of Leave-One-Channel-Out Testing.} We present the detailed results for column "7" in Table~\ref{table:jumpcp8_partial} of the main paper. Each row shows the results obtained by leaving out one channel and testing the model on the remaining seven channels. For all experiments conducted, we provide the \textit{mean} $\pm$ \textit{standard deviation} based on three runs. Our method, \ours{}, consistently outperforms the best two baselines \channelvit~\cite{bao2024channel} and \dichavit~\cite{phamDiChaVit2024} by $17-23\%$ across all tested combinations.}
\label{table:leave_one_channel_out}
\setlength{\tabcolsep}{6pt}
\centering
\begin{tabular}{cccc}
    \toprule
    Missing channel at inference & \channelvit~\cite{bao2024channel} & \dichavit~\cite{phamDiChaVit2024} & \textbf{\oursbr} \\
    \midrule
    $0$ & 61.72 $\pm$ 0.48 & {63.48 $\pm$ 0.20} & \textbf{86.29} $\pm$ 0.37  \\
    $1$ & 61.21 $\pm$ 0.41 & {62.72 $\pm$ 0.28} & \textbf{86.16} $\pm$ 0.32  \\
    $2$ & 61.90 $\pm$ 0.48  & {63.28 $\pm$ 0.31} & \textbf{87.13} $\pm$ 0.24  \\
    $3$ & 37.70 $\pm$ 0.60  & {38.83 $\pm$ 0.46} & \textbf{61.79} $\pm$ 0.58  \\
    $4$ & 58.52 $\pm$ 0.63 & {59.61 $\pm$ 0.17} & \textbf{82.67} $\pm$ 0.53 \\
    $5$ & 67.28 $\pm$ 0.53 & {69.12 $\pm$ 0.16} & \textbf{89.24} $\pm$ 0.35 \\
    $6$ & 67.20 $\pm$ 0.59  & {69.06 $\pm$ 0.20}& \textbf{87.54} $\pm$ 0.20 \\
    $7$ & 67.37 $\pm$ 0.60 & {69.21 $\pm$ 0.19} & \textbf{86.03} $\pm$ 0.32 \\
    \bottomrule
\end{tabular}
\end{table*}

%% file: tables/bigger_model.tex
\begin{table*}[t]
\caption{\textbf{Scaling Model Size.} We compare \ours{} with the two best baseline models, \channelvit~\cite{bao2024channel} and \dichavit~\cite{phamDiChaVit2024}. Each model is evaluated using two backbones: ViT-S ($21$M parameters) and ViT-B ($85$M parameters). Increasing the model size boosts performance, particularly on CHAMMI and JUMP-CP. Our method outperforms others across three datasets with both backbones.}
\label{tab:scale_bigger}
\centering
\begin{tabular}{r*{7}{c}}

\toprule
& & \multicolumn{1}{c}{CHAMMI~\cite{chen2023chammi}} & \multicolumn{2}{c}{JUMP-CP~\cite{jumpcpSrinivas}} & \multicolumn{2}{c}{So2Sat~\cite{mediatum1483140}} \\
\cmidrule(r){3-3} \cmidrule(r){4-5} \cmidrule(r){6-7}
Model & Backbone & Avg score & Full & Partial & Full & Partial \\
\midrule

   \channelvit~\cite{bao2024channel} & ViT-S & 64.97 & 67.51 & 56.49 & 61.03 & 46.16 \\

\dichavit~\cite{phamDiChaVit2024} & ViT-S & 69.77 & 69.19 & 57.98 & 63.36 & 47.76 \\

\textbf{\oursbr} & ViT-S & 74.63 & 90.73 & 68.05 & 67.44 & 52.11 \\

 \midrule    

\channelvit~\cite{bao2024channel} & ViT-B & 67.72 & 67.92 & 56.97 & 62.19 & 46.20 \\

\dichavit~\cite{phamDiChaVit2024} & ViT-B & 70.56  & 69.47 & 58.33 & 63.93 & 47.92 \\


\textbf{\oursbr} & ViT-B & \textbf{77.09} & \textbf{92.22} & \textbf{70.24} & \textbf{67.67} & \textbf{52.27}\\

\bottomrule
\end{tabular}

\end{table*}

%% file: 0_main.bbl
\begin{thebibliography}{63}
\providecommand{\natexlab}[1]{#1}
\providecommand{\url}[1]{\texttt{#1}}
\expandafter\ifx\csname urlstyle\endcsname\relax
  \providecommand{\doi}[1]{doi: #1}\else
  \providecommand{\doi}{doi: \begingroup \urlstyle{rm}\Url}\fi

\bibitem[Kraus et~al.(2024)Kraus, Kenyon-Dean, Saberian, Fallah, McLean, Leung, Sharma, Khan, Balakrishnan, Celik, et~al.]{kraus2024masked}
Oren Kraus, Kian Kenyon-Dean, Saber Saberian, Maryam Fallah, Peter McLean, Jess Leung, Vasudev Sharma, Ayla Khan, Jia Balakrishnan, Safiye Celik, et~al.
\newblock Masked autoencoders for microscopy are scalable learners of cellular biology.
\newblock In \emph{Proceedings of the IEEE/CVF Conference on Computer Vision and Pattern Recognition}, pages 11757--11768, 2024.

\bibitem[Szegedy et~al.(2015)Szegedy, Liu, Jia, Sermanet, and Reed]{inception}
Christian Szegedy, Wei Liu, Yangqing Jia, Pierre Sermanet, and Scott Reed.
\newblock Dragomiranguelov, dumitru erhan, vincent vanhoucke, and andrew rabinovich. 2015. going deeper with convolutions.
\newblock In \emph{Proceedings of the IEEE conference on computer vision and pattern recognition}, pages 1--9, 2015.

\bibitem[Li et~al.(2017)Li, Ding, Wang, and Cao]{resnet2}
Xuelei Li, Liangkui Ding, Li~Wang, and Fang Cao.
\newblock Fpga accelerates deep residual learning for image recognition.
\newblock In \emph{2017 IEEE 2nd Information Technology, Networking, Electronic and Automation Control Conference (ITNEC)}, pages 837--840. IEEE, 2017.

\bibitem[Tan and Le(2019)]{efficientnet}
Mingxing Tan and Quoc Le.
\newblock Efficientnet: Rethinking model scaling for convolutional neural networks.
\newblock In \emph{International conference on machine learning}, pages 6105--6114. PMLR, 2019.

\bibitem[Howard et~al.(2017)Howard, Zhu, Chen, Kalenichenko, Wang, Weyand, Andreetto, and Adam]{mobilenet}
Andrew~G Howard, Menglong Zhu, Bo~Chen, Dmitry Kalenichenko, Weijun Wang, Tobias Weyand, Marco Andreetto, and Hartwig Adam.
\newblock Mobilenets: Efficient convolutional neural networks for mobile vision applications.
\newblock \emph{arXiv preprint arXiv:1704.04861}, 2017.

\bibitem[Liu et~al.(2021)Liu, Lin, Cao, Hu, Wei, Zhang, Lin, and Guo]{swin}
Ze~Liu, Yutong Lin, Yue Cao, Han Hu, Yixuan Wei, Zheng Zhang, Stephen Lin, and Baining Guo.
\newblock Swin transformer: Hierarchical vision transformer using shifted windows.
\newblock In \emph{Proceedings of the IEEE/CVF international conference on computer vision}, pages 10012--10022, 2021.

\bibitem[Liu et~al.(2022)Liu, Mao, Wu, Feichtenhofer, Darrell, and Xie]{liu2022convnet}
Zhuang Liu, Hanzi Mao, Chao-Yuan Wu, Christoph Feichtenhofer, Trevor Darrell, and Saining Xie.
\newblock A convnet for the 2020s.
\newblock In \emph{IEEE/CVF Conference on Computer Vision and Pattern Recognition (CVPR)}, 2022.

\bibitem[Ryali et~al.(2023)Ryali, Hu, Bolya, Wei, Fan, Huang, Aggarwal, Chowdhury, Poursaeed, Hoffman, et~al.]{hiera}
Chaitanya Ryali, Yuan-Ting Hu, Daniel Bolya, Chen Wei, Haoqi Fan, Po-Yao Huang, Vaibhav Aggarwal, Arkabandhu Chowdhury, Omid Poursaeed, Judy Hoffman, et~al.
\newblock Hiera: A hierarchical vision transformer without the bells-and-whistles.
\newblock \emph{arXiv preprint arXiv:2306.00989}, 2023.

\bibitem[Liu et~al.(2025)Liu, Tian, Zhao, Yu, Xie, Wang, Ye, Jiao, and Liu]{liu2025vmamba}
Yue Liu, Yunjie Tian, Yuzhong Zhao, Hongtian Yu, Lingxi Xie, Yaowei Wang, Qixiang Ye, Jianbin Jiao, and Yunfan Liu.
\newblock Vmamba: Visual state space model.
\newblock \emph{Advances in neural information processing systems}, 37:\penalty0 103031--103063, 2025.

\bibitem[Zhu et~al.(2019)Zhu, Hu, Qiu, Shi, Bagheri, Kang, Li, Mou, Zhang, Häberle, Han, Hua, Huang, Hughes, Sun, Schmitt, and Wang]{mediatum1483140}
Xiaoxiang Zhu, Jingliang Hu, Chunping Qiu, Yilei Shi, Hossein Bagheri, Jian Kang, Hao Li, Lichao Mou, Guicheng Zhang, Matthias Häberle, Shiyao Han, Yuansheng Hua, Rong Huang, Lloyd Hughes, Yao Sun, Michael Schmitt, and Yuanyuan Wang.
\newblock New: So2sat lcz42, 2019.
\newblock URL \url{https://mediatum.ub.tum.de/1483140}.

\bibitem[Zou et~al.(2024)Zou, Huang, Yang, Guo, Luo, Feng, and Zuo]{zou2024unim}
Jian Zou, Tianyu Huang, Guanglei Yang, Zhenhua Guo, Tao Luo, Chun-Mei Feng, and Wangmeng Zuo.
\newblock Unim 2 ae: Multi-modal masked autoencoders with unified 3d representation for 3d perception in autonomous driving.
\newblock In \emph{European Conference on Computer Vision}, pages 296--313. Springer, 2024.

\bibitem[Chen et~al.(2023)Chen, Pham, Wang, Doron, Moshkov, Plummer, and Caicedo]{chen2023chammi}
Zitong Chen, Chau Pham, Siqi Wang, Michael Doron, Nikita Moshkov, Bryan~A. Plummer, and Juan~C Caicedo.
\newblock {CHAMMI}: A benchmark for channel-adaptive models in microscopy imaging.
\newblock In \emph{Thirty-seventh Conference on Neural Information Processing Systems Datasets and Benchmarks Track}, 2023.
\newblock URL \url{https://openreview.net/forum?id=Luc1bZLeMY}.

\bibitem[Bourriez et~al.(2024)Bourriez, Bendidi, Ethan, Watkinson, Sanchez, Bollot, and Genovesio]{chadavit}
Nicolas Bourriez, Ihab Bendidi, Cohen Ethan, Gabriel Watkinson, Maxime Sanchez, Guillaume Bollot, and Auguste Genovesio.
\newblock Chada-vit : Channel adaptive attention for joint representation learning of heterogeneous microscopy images.
\newblock In \emph{The IEEE Conference on Computer Vision and Pattern Recognition (CVPR)}, 2024.

\bibitem[Chen et~al.(2017)Chen, Carass, Jog, Lee, Roy, and Prince]{chen2017cross}
Min Chen, Aaron Carass, Amod Jog, Junghoon Lee, Snehashis Roy, and Jerry~L Prince.
\newblock Cross contrast multi-channel image registration using image synthesis for mr brain images.
\newblock \emph{Medical image analysis}, 36:\penalty0 2--14, 2017.

\bibitem[Kenyon-Dean et~al.(2025{\natexlab{a}})Kenyon-Dean, Wang, Urbanik, Donhauser, Hartford, Saberian, Sahin, Bendidi, Celik, Vera, Fay, Haque, and Kraus]{kenyon-dean2025vitally}
Kian Kenyon-Dean, Zitong~Jerry Wang, John Urbanik, Konstantin Donhauser, Jason Hartford, Saber Saberian, Nil Sahin, Ihab Bendidi, Safiye Celik, Juan Sebasti{\'a}n~Rodr{\'\i}guez Vera, Marta Fay, Imran~S Haque, and Oren Kraus.
\newblock Vitally consistent: Scaling biological representation learning for cell microscopy, 2025{\natexlab{a}}.
\newblock URL \url{https://openreview.net/forum?id=niywLsa54R}.

\bibitem[Darcet et~al.(2024)Darcet, Oquab, Mairal, and Bojanowski]{darcet2024vision}
Timoth{\'e}e Darcet, Maxime Oquab, Julien Mairal, and Piotr Bojanowski.
\newblock Vision transformers need registers.
\newblock In \emph{The Twelfth International Conference on Learning Representations}, 2024.
\newblock URL \url{https://openreview.net/forum?id=2dnO3LLiJ1}.

\bibitem[Bachmann et~al.(2022)Bachmann, Mizrahi, Atanov, and Zamir]{bachmann2022multimae}
Roman Bachmann, David Mizrahi, Andrei Atanov, and Amir Zamir.
\newblock Multimae: Multi-modal multi-task masked autoencoders.
\newblock In \emph{European Conference on Computer Vision}, pages 348--367. Springer, 2022.

\bibitem[Zhang et~al.(2024)Zhang, Yang, Wang, Li, and Yin]{zhang2024multimodal}
Xiang Zhang, Huiyuan Yang, Taoyue Wang, Xiaotian Li, and Lijun Yin.
\newblock Multimodal channel-mixing: Channel and spatial masked autoencoder on facial action unit detection.
\newblock In \emph{Proceedings of the IEEE/CVF Winter Conference on Applications of Computer Vision}, pages 6077--6086, 2024.

\bibitem[Liang et~al.(2022)Liang, Li, and Marculescu]{liang2022supmae}
Feng Liang, Yangguang Li, and Diana Marculescu.
\newblock Supmae: Supervised masked autoencoders are efficient vision learners.
\newblock \emph{arXiv preprint arXiv:2205.14540}, 2022.

\bibitem[Cheng et~al.(2021)Cheng, Wang, Jiang, and Macherey]{cheng2021self}
Yong Cheng, Wei Wang, Lu~Jiang, and Wolfgang Macherey.
\newblock Self-supervised and supervised joint training for resource-rich machine translation.
\newblock In \emph{International Conference on Machine Learning}, pages 1825--1835. PMLR, 2021.

\bibitem[Bai et~al.(2022)Bai, Li, Zhang, Bapna, Siddhartha, Sim, and Sainath]{bai2022joint}
Junwen Bai, Bo~Li, Yu~Zhang, Ankur Bapna, Nikhil Siddhartha, Khe~Chai Sim, and Tara~N Sainath.
\newblock Joint unsupervised and supervised training for multilingual asr.
\newblock In \emph{ICASSP 2022-2022 IEEE International Conference on Acoustics, Speech and Signal Processing (ICASSP)}, pages 6402--6406. IEEE, 2022.

\bibitem[Xin et~al.(2023)Xin, Peng, and Lu]{xin2024masked}
Yifei Xin, Xiulian Peng, and Yan Lu.
\newblock Masked audio modeling with clap and multi-objective learning.
\newblock In \emph{INTERSPEECH}, pages 2763--2767, 2023.
\newblock URL \url{https://doi.org/10.21437/Interspeech.2023-2488}.

\bibitem[Gao et~al.(2020)Gao, Han, Wang, Huang, and Scott]{gao2020channel}
Yu~Gao, Xintong Han, Xun Wang, Weilin Huang, and Matthew Scott.
\newblock Channel interaction networks for fine-grained image categorization.
\newblock In \emph{Proceedings of the AAAI conference on artificial intelligence}, volume~34, pages 10818--10825, 2020.

\bibitem[Zhou et~al.(2022{\natexlab{a}})Zhou, Yu, Xie, Xiao, Anandkumar, Feng, and Alvarez]{zhou2022understanding}
Daquan Zhou, Zhiding Yu, Enze Xie, Chaowei Xiao, Animashree Anandkumar, Jiashi Feng, and Jose~M Alvarez.
\newblock Understanding the robustness in vision transformers.
\newblock In \emph{International Conference on Machine Learning}, pages 27378--27394. PMLR, 2022{\natexlab{a}}.

\bibitem[Hu et~al.(2018)Hu, Shen, and Sun]{hu2018squeeze}
Jie Hu, Li~Shen, and Gang Sun.
\newblock Squeeze-and-excitation networks.
\newblock In \emph{Proceedings of the IEEE conference on computer vision and pattern recognition}, pages 7132--7141, 2018.

\bibitem[Yang et~al.(2019)Yang, Ren, Gan, Zhu, and Parikh]{yang2019cross}
Jianwei Yang, Zhile Ren, Chuang Gan, Hongyuan Zhu, and Devi Parikh.
\newblock Cross-channel communication networks.
\newblock \emph{Advances in Neural Information Processing Systems}, 32, 2019.

\bibitem[Chandrasekaran et~al.(2022)Chandrasekaran, Cimini, Goodale, Miller, Kost-Alimova, Jamali, Doench, Fritchman, Skepner, Melanson, Arevalo, Caicedo, Kuhn, Hernandez, Berstler, Shafqat-Abbasi, Root, Swalley, Singh, and Carpenter]{jumpcpSrinivas}
Srinivas~Niranj Chandrasekaran, Beth~A. Cimini, Amy Goodale, Lisa Miller, Maria Kost-Alimova, Nasim Jamali, John Doench, Briana Fritchman, Adam Skepner, Michelle Melanson, John Arevalo, Juan~C. Caicedo, Daniel Kuhn, Desiree Hernandez, Jim Berstler, Hamdah Shafqat-Abbasi, David Root, Sussane Swalley, Shantanu Singh, and Anne~E. Carpenter.
\newblock Three million images and morphological profiles of cells treated with matched chemical and genetic perturbations.
\newblock \emph{bioRxiv}, 2022.
\newblock \doi{10.1101/2022.01.05.475090}.
\newblock URL \url{https://www.biorxiv.org/content/early/2022/01/05/2022.01.05.475090}.

\bibitem[Mohajerani and Saeedi(2019)]{mohajerani2019cloud}
Sorour Mohajerani and Parvaneh Saeedi.
\newblock Cloud-net: An end-to-end cloud detection algorithm for landsat 8 imagery.
\newblock In \emph{IGARSS 2019-2019 IEEE international geoscience and remote sensing symposium}, pages 1029--1032. IEEE, 2019.

\bibitem[Pham and Plummer(2024)]{phamDiChaVit2024}
Chau Pham and Bryan~A. Plummer.
\newblock Enhancing feature diversity boosts channel-adaptive vision transformers.
\newblock In \emph{Advances in Neural Information Processing Systems (NeurIPS)}, 2024.

\bibitem[Bao et~al.(2024)Bao, Sivanandan, and Karaletsos]{bao2024channel}
Yujia Bao, Srinivasan Sivanandan, and Theofanis Karaletsos.
\newblock Channel vision transformers: An image is worth 1 x 16 x 16 words.
\newblock In \emph{The Twelfth International Conference on Learning Representations}, 2024.
\newblock URL \url{https://openreview.net/forum?id=CK5Hfb5hBG}.

\bibitem[He et~al.(2022)He, Chen, Xie, Li, Doll{\'a}r, and Girshick]{he2022masked}
Kaiming He, Xinlei Chen, Saining Xie, Yanghao Li, Piotr Doll{\'a}r, and Ross Girshick.
\newblock Masked autoencoders are scalable vision learners.
\newblock In \emph{Proceedings of the IEEE/CVF conference on computer vision and pattern recognition}, pages 16000--16009, 2022.

\bibitem[Xie et~al.(2022)Xie, Zhang, Cao, Lin, Bao, Yao, Dai, and Hu]{xie2022simmim}
Zhenda Xie, Zheng Zhang, Yue Cao, Yutong Lin, Jianmin Bao, Zhuliang Yao, Qi~Dai, and Han Hu.
\newblock Simmim: A simple framework for masked image modeling.
\newblock In \emph{Proceedings of the IEEE/CVF conference on computer vision and pattern recognition}, pages 9653--9663, 2022.

\bibitem[Sims et~al.(2023)Sims, Mills, and Chang]{sims2023mim}
Zachary Sims, Gordon~B Mills, and Young~Hwan Chang.
\newblock Mim-cycif: masked imaging modeling for enhancing cyclic immunofluorescence (cycif) with panel reduction and imputation.
\newblock \emph{bioRxiv}, 2023.

\bibitem[Lin et~al.(2023)Lin, Gao, Shi, Dong, and Du]{lin2023ss}
Junyan Lin, Feng Gao, Xiaochen Shi, Junyu Dong, and Qian Du.
\newblock Ss-mae: Spatial--spectral masked autoencoder for multisource remote sensing image classification.
\newblock \emph{IEEE Transactions on Geoscience and Remote Sensing}, 61:\penalty0 1--14, 2023.

\bibitem[Wang et~al.(2024)Wang, Wen, Zhang, Sun, Yang, Zhang, and Lu]{wang2024hsimae}
Yue Wang, Ming Wen, Hailiang Zhang, Jinyu Sun, Qiong Yang, Zhimin Zhang, and Hongmei Lu.
\newblock Hsimae: A unified masked autoencoder with large-scale pre-training for hyperspectral image classification.
\newblock \emph{IEEE Journal of Selected Topics in Applied Earth Observations and Remote Sensing}, 2024.

\bibitem[Geiping et~al.(2023)Geiping, Garrido, Fernandez, Bar, Pirsiavash, LeCun, and Goldblum]{geiping2023cookbook}
Jonas Geiping, Quentin Garrido, Pierre Fernandez, Amir Bar, Hamed Pirsiavash, Yann LeCun, and Micah Goldblum.
\newblock A cookbook of self-supervised learning.
\newblock \emph{arXiv preprint arXiv:2304.12210}, 2023.

\bibitem[Chen et~al.(2020{\natexlab{a}})Chen, Kornblith, Norouzi, and Hinton]{chen2020simple}
Ting Chen, Simon Kornblith, Mohammad Norouzi, and Geoffrey Hinton.
\newblock A simple framework for contrastive learning of visual representations.
\newblock In \emph{International conference on machine learning}, pages 1597--1607. PMLR, 2020{\natexlab{a}}.

\bibitem[Chen et~al.(2020{\natexlab{b}})Chen, Kornblith, Swersky, Norouzi, and Hinton]{chen2020big}
Ting Chen, Simon Kornblith, Kevin Swersky, Mohammad Norouzi, and Geoffrey~E Hinton.
\newblock Big self-supervised models are strong semi-supervised learners.
\newblock \emph{Advances in neural information processing systems}, 33:\penalty0 22243--22255, 2020{\natexlab{b}}.

\bibitem[Dwibedi et~al.(2021)Dwibedi, Aytar, Tompson, Sermanet, and Zisserman]{dwibedi2021little}
Debidatta Dwibedi, Yusuf Aytar, Jonathan Tompson, Pierre Sermanet, and Andrew Zisserman.
\newblock With a little help from my friends: Nearest-neighbor contrastive learning of visual representations.
\newblock In \emph{Proceedings of the IEEE/CVF International Conference on Computer Vision}, pages 9588--9597, 2021.

\bibitem[Chen and He(2021)]{chen2021exploring}
Xinlei Chen and Kaiming He.
\newblock Exploring simple siamese representation learning.
\newblock In \emph{Proceedings of the IEEE/CVF conference on computer vision and pattern recognition}, pages 15750--15758, 2021.

\bibitem[Grill et~al.(2020)Grill, Strub, Altch{\'e}, Tallec, Richemond, Buchatskaya, Doersch, Avila~Pires, Guo, Gheshlaghi~Azar, et~al.]{grill2020bootstrap}
Jean-Bastien Grill, Florian Strub, Florent Altch{\'e}, Corentin Tallec, Pierre Richemond, Elena Buchatskaya, Carl Doersch, Bernardo Avila~Pires, Zhaohan Guo, Mohammad Gheshlaghi~Azar, et~al.
\newblock Bootstrap your own latent-a new approach to self-supervised learning.
\newblock \emph{Advances in neural information processing systems}, 33:\penalty0 21271--21284, 2020.

\bibitem[Caron et~al.(2021)Caron, Touvron, Misra, J{\'e}gou, Mairal, Bojanowski, and Joulin]{caron2021emerging}
Mathilde Caron, Hugo Touvron, Ishan Misra, Herv{\'e} J{\'e}gou, Julien Mairal, Piotr Bojanowski, and Armand Joulin.
\newblock Emerging properties in self-supervised vision transformers.
\newblock In \emph{Proceedings of the IEEE/CVF international conference on computer vision}, pages 9650--9660, 2021.

\bibitem[Zhou et~al.(2022{\natexlab{b}})Zhou, Wei, Wang, Shen, Xie, Yuille, and Kong]{zhou2021ibot}
Jinghao Zhou, Chen Wei, Huiyu Wang, Wei Shen, Cihang Xie, Alan Yuille, and Tao Kong.
\newblock ibot: Image bert pre-training with online tokenizer.
\newblock \emph{International Conference on Learning Representations (ICLR)}, 2022{\natexlab{b}}.

\bibitem[Oquab et~al.(2024)Oquab, Darcet, Moutakanni, Vo, Szafraniec, Khalidov, Fernandez, HAZIZA, Massa, El-Nouby, Assran, Ballas, Galuba, Howes, Huang, Li, Misra, Rabbat, Sharma, Synnaeve, Xu, Jegou, Mairal, Labatut, Joulin, and Bojanowski]{oquab2024dinov}
Maxime Oquab, Timoth{\'e}e Darcet, Th{\'e}o Moutakanni, Huy~V. Vo, Marc Szafraniec, Vasil Khalidov, Pierre Fernandez, Daniel HAZIZA, Francisco Massa, Alaaeldin El-Nouby, Mido Assran, Nicolas Ballas, Wojciech Galuba, Russell Howes, Po-Yao Huang, Shang-Wen Li, Ishan Misra, Michael Rabbat, Vasu Sharma, Gabriel Synnaeve, Hu~Xu, Herve Jegou, Julien Mairal, Patrick Labatut, Armand Joulin, and Piotr Bojanowski.
\newblock {DINO}v2: Learning robust visual features without supervision.
\newblock \emph{Transactions on Machine Learning Research}, 2024.
\newblock ISSN 2835-8856.
\newblock URL \url{https://openreview.net/forum?id=a68SUt6zFt}.
\newblock Featured Certification.

\bibitem[Assran et~al.(2022)Assran, Caron, Misra, Bojanowski, Bordes, Vincent, Joulin, Rabbat, and Ballas]{assran2022masked}
Mahmoud Assran, Mathilde Caron, Ishan Misra, Piotr Bojanowski, Florian Bordes, Pascal Vincent, Armand Joulin, Mike Rabbat, and Nicolas Ballas.
\newblock Masked siamese networks for label-efficient learning.
\newblock In \emph{European Conference on Computer Vision}, pages 456--473. Springer, 2022.

\bibitem[Roddan et~al.(2024)Roddan, Czempiel, Elson, and Giannarou]{roddan2024calibration}
Alfie Roddan, Tobias Czempiel, Daniel~S Elson, and Stamatia Giannarou.
\newblock Calibration-jitter: Augmentation of hyperspectral data for improved surgical scene segmentation.
\newblock \emph{Healthcare Technology Letters}, 11\penalty0 (6):\penalty0 345--354, 2024.

\bibitem[Patnala et~al.(2023)Patnala, Stadtler, Schultz, and Gall]{patnala2023generating}
Ankit Patnala, Scarlet Stadtler, Martin~G Schultz, and Juergen Gall.
\newblock Generating views using atmospheric correction for contrastive self-supervised learning of multispectral images.
\newblock \emph{IEEE Geoscience and Remote Sensing Letters}, 20:\penalty0 1--5, 2023.

\bibitem[Nedungadi et~al.(2024)Nedungadi, Kariryaa, Oehmcke, Belongie, Igel, and Lang]{nedungadi2024mmearth}
Vishal Nedungadi, Ankit Kariryaa, Stefan Oehmcke, Serge Belongie, Christian Igel, and Nico Lang.
\newblock Mmearth: Exploring multi-modal pretext tasks for geospatial representation learning.
\newblock In \emph{European Conference on Computer Vision}, pages 164--182. Springer, 2024.

\bibitem[Oquab et~al.(2023)Oquab, Darcet, Moutakanni, Vo, Szafraniec, Khalidov, Fernandez, Haziza, Massa, El-Nouby, Howes, Huang, Xu, Sharma, Li, Galuba, Rabbat, Assran, Ballas, Synnaeve, Misra, Jegou, Mairal, Labatut, Joulin, and Bojanowski]{oquab2023dinov2}
Maxime Oquab, Timothée Darcet, Theo Moutakanni, Huy~V. Vo, Marc Szafraniec, Vasil Khalidov, Pierre Fernandez, Daniel Haziza, Francisco Massa, Alaaeldin El-Nouby, Russell Howes, Po-Yao Huang, Hu~Xu, Vasu Sharma, Shang-Wen Li, Wojciech Galuba, Mike Rabbat, Mido Assran, Nicolas Ballas, Gabriel Synnaeve, Ishan Misra, Herve Jegou, Julien Mairal, Patrick Labatut, Armand Joulin, and Piotr Bojanowski.
\newblock Dinov2: Learning robust visual features without supervision, 2023.

\bibitem[Hendrycks and Gimpel(2016)]{hendrycks2016gaussian}
Dan Hendrycks and Kevin Gimpel.
\newblock Gaussian error linear units (gelus).
\newblock \emph{arXiv preprint arXiv:1606.08415}, 2016.

\bibitem[Ha et~al.(2016)Ha, Dai, and Le]{haHypernetworks2016}
David Ha, Andrew Dai, and Quoc Le.
\newblock Hypernetworks.
\newblock In \emph{International Conference on Learning Representations (ICLR)}, 2016.

\bibitem[Plummer et~al.(2022)Plummer, Dryden, Frost, Hoefler, and Saenko]{plummerNPAS2022}
Bryan~A. Plummer, Nikoli Dryden, Julius Frost, Torsten Hoefler, and Kate Saenko.
\newblock Neural parameter allocation search.
\newblock In \emph{International Conference on Learning Representations (ICLR)}, 2022.

\bibitem[Savarese and Maire(2019)]{savarese2018learning}
Pedro Savarese and Michael Maire.
\newblock Learning implicitly recurrent {CNN}s through parameter sharing.
\newblock In \emph{International Conference on Learning Representations (ICLR)}, 2019.

\bibitem[Pham et~al.(2024)Pham, Teterwak, Nelson, and Plummer]{phamMixtureGrowth2024}
Chau Pham, Piotr Teterwak, Soren Nelson, and Bryan~A. Plummer.
\newblock Mixturegrowth: Growing neural networks by recombining learned parameters.
\newblock In \emph{IEEE Winter Conference on Applications of Computer Vision (WACV)}, 2024.

\bibitem[Kenyon-Dean et~al.(2025{\natexlab{b}})Kenyon-Dean, Wang, Urbanik, Donhauser, Hartford, Saberian, Sahin, Bendidi, Celik, Vera, Fay, Haque, and Kraus]{kenyon2024vitally}
Kian Kenyon-Dean, Zitong~Jerry Wang, John Urbanik, Konstantin Donhauser, Jason Hartford, Saber Saberian, Nil Sahin, Ihab Bendidi, Safiye Celik, Juan Sebasti{\'a}n~Rodr{\'\i}guez Vera, Marta Fay, Imran~S Haque, and Oren Kraus.
\newblock Vitally consistent: Scaling biological representation learning for cell microscopy, 2025{\natexlab{b}}.
\newblock URL \url{https://openreview.net/forum?id=niywLsa54R}.

\bibitem[Eyma{\"e}l et~al.(2024)Eyma{\"e}l, Vandeghen, Cioppa, Giancola, Ghanem, and Van~Droogenbroeck]{eymael2024efficient}
Alexandre Eyma{\"e}l, Renaud Vandeghen, Anthony Cioppa, Silvio Giancola, Bernard Ghanem, and Marc Van~Droogenbroeck.
\newblock Efficient image pre-training with siamese cropped masked autoencoders.
\newblock In \emph{European Conference on Computer Vision}, pages 348--366. Springer, 2024.

\bibitem[Bao et~al.(2022)Bao, Dong, Piao, and Wei]{bao2022beit}
Hangbo Bao, Li~Dong, Songhao Piao, and Furu Wei.
\newblock {BE}it: {BERT} pre-training of image transformers.
\newblock In \emph{International Conference on Learning Representations}, 2022.
\newblock URL \url{https://openreview.net/forum?id=p-BhZSz59o4}.

\bibitem[Sablayrolles et~al.(2019)Sablayrolles, Douze, Schmid, and Jégou]{sablayrolles2018spreading}
Alexandre Sablayrolles, Matthijs Douze, Cordelia Schmid, and Hervé Jégou.
\newblock Spreading vectors for similarity search.
\newblock In \emph{International Conference on Learning Representations}, 2019.
\newblock URL \url{https://openreview.net/forum?id=SkGuG2R5tm}.

\bibitem[Donato and Belongie(2002)]{donato2002approximate}
Gianluca Donato and Serge Belongie.
\newblock Approximate thin plate spline mappings.
\newblock In \emph{Computer Vision—ECCV 2002: 7th European Conference on Computer Vision Copenhagen, Denmark, May 28--31, 2002 Proceedings, Part III 7}, pages 21--31. Springer, 2002.

\bibitem[Teh et~al.(2020)Teh, DeVries, and Taylor]{teh2020proxynca}
Eu~Wern Teh, Terrance DeVries, and Graham~W Taylor.
\newblock Proxynca++: Revisiting and revitalizing proxy neighborhood component analysis.
\newblock In \emph{Computer Vision--ECCV 2020: 16th European Conference, Glasgow, UK, August 23--28, 2020, Proceedings, Part XXIV 16}, pages 448--464. Springer, 2020.

\bibitem[Noman et~al.(2024)Noman, Naseer, Cholakkal, Anwer, Khan, and Khan]{noman2024rethinking}
Mubashir Noman, Muzammal Naseer, Hisham Cholakkal, Rao~Muhammad Anwer, Salman Khan, and Fahad~Shahbaz Khan.
\newblock Rethinking transformers pre-training for multi-spectral satellite imagery.
\newblock In \emph{Proceedings of the IEEE/CVF Conference on Computer Vision and Pattern Recognition}, pages 27811--27819, 2024.

\bibitem[Reed et~al.(2022)Reed, Gupta, Li, Brockman, Funk, Clipp, Candido, UyttenDAele, and Darrell]{reed2022scale}
C~Reed, Ritwik Gupta, Shufan Li, Sarah Brockman, Christopher Funk, Brian Clipp, S~Candido, M~UyttenDAele, and T~Darrell.
\newblock Scale-mae: A scale-aware masked autoencoder for multiscale geospatial representation learning. 2023 ieee.
\newblock In \emph{CVF International Conference on Computer Vision (ICCV)}, pages 4065--4076, 2022.

\bibitem[Mustafa et~al.(2022)Mustafa, Mikhailiuk, Iliescu, Babbar, and Mantiuk]{mustafa2022training}
Aamir Mustafa, Aliaksei Mikhailiuk, Dan~Andrei Iliescu, Varun Babbar, and Rafa{\l}~K Mantiuk.
\newblock Training a task-specific image reconstruction loss.
\newblock In \emph{Proceedings of the IEEE/CVF winter conference on applications of computer vision}, pages 2319--2328, 2022.

\end{thebibliography}
